\documentclass[fleqn,11pt]{wlscirep}
\usepackage[utf8]{inputenc}
\usepackage[T1]{fontenc}
\usepackage{booktabs}
\usepackage{rotating}
\usepackage{marvosym}
\usepackage{lineno}
\usepackage{hyperref}
\usepackage{multirow}
\usepackage{graphicx} %
\usepackage{adjustbox}
\usepackage{makecell}
\usepackage{xcolor}  

\newcommand{\modelname}{InstructBioMol}

\title{Advancing Biomolecular Understanding and Design Following Human Instructions}

\author[1,2]{Xiang Zhuang}
\author[2\Letter]{Keyan Ding}
\author[2,5]{Tianwen Lyu}
\author[1,2]{Yinuo Jiang}
\author[1,2]{Xiaotong Li}
\author[2,5]{Zhuoyi Xiang}
\author[1,2]{Zeyuan Wang}
\author[2,6]{Ming Qin}
\author[1,2]{Kehua Feng}
\author[7]{Jike Wang}
\author[3,2\Letter]{Qiang Zhang}
\author[1,2,4\Letter]{Huajun Chen}
\affil[1]{College of Computer Science and Technology, Zhejiang University, Hangzhou, China}
\affil[2]{Zhejiang Key Laboratory of Intelligent Manufacturing for Functional Chemicals, ZJU-Hangzhou Global Scientific and Technological Innovation Center, Zhejiang University, Hangzhou, China}
\affil[3]{ZJU-UIUC Institute, International Campus, Zhejiang University, Haining, China}
\affil[4]{State Key Laboratory of Ocean Sensing, Hangzhou,
China}
\affil[5]{Polytechnic Institute, Zhejiang University, Hangzhou,
China}
\affil[6]{School of Software Technology, Zhejiang University, Hangzhou, China}
\affil[7]{College of Pharmaceutical Sciences, Zhejiang University, Hangzhou, China}

\affil[\Letter]{Corresponding author}

\begin{abstract}
    Understanding and designing biomolecules, such as proteins and small molecules, is central to advancing drug discovery, synthetic biology, and enzyme engineering. 
    Recent breakthroughs in Artificial Intelligence (AI) have revolutionized biomolecular research,
    achieving remarkable accuracy in biomolecular prediction and design.
    {However, a critical gap remains between AI’s computational capabilities and researchers’ intuitive goals, particularly in using natural language to bridge complex tasks with human intentions.}
    Large Language Models (LLMs) have shown potential to interpret human intentions, yet their application to biomolecular research remains nascent due to challenges including specialized knowledge requirements, multimodal data integration, and semantic alignment between natural language and biomolecules. To address these limitations, we present \modelname{}, a novel LLM designed to bridge natural language and biomolecules through a comprehensive any-to-any alignment of natural language, molecules, and proteins.
    {This model can integrate multimodal biomolecules as input, and enable researchers to articulate design goals in natural language, providing biomolecular outputs that meet precise biological needs.} {Experimental results demonstrate \modelname{} can understand and design biomolecules following human instructions.
    Notably, it can generate drug molecules with a 10\% improvement in binding affinity and design enzymes
    that achieve an {Enzyme-Substrate Pair Prediction Score (ESP Score)} of 70.4, making it the only method to surpass the enzyme-substrate interaction threshold of 60.0 recommended by the ESP developer. This highlights its potential to transform real-world biomolecular research.}
\end{abstract}
\begin{document}
\flushbottom
\maketitle
%
%
\thispagestyle{empty}

\section*{Introduction}

Understanding and designing biomolecules is fundamental to natural science research. Biomolecules, such as proteins and small molecules, play essential roles in biological processes, and their precise manipulation is key to advancements in drug discovery, synthetic biology, and enzyme engineering~\cite{kim2021comprehensive,volk2020biosystems,mazurenko2019machine}. Recent Artificial Intelligence (AI) breakthroughs have transformed research in these areas~\cite{abramson2024accurate,krishna2024generalized}. Tools like AlphaFold3~\cite{abramson2024accurate} and RoseTTAFold All-Atom~\cite{krishna2024generalized} have revolutionized biomolecular structure prediction, offering unprecedented accuracy and speed.
{Despite these advancements, a crucial challenge persists: how to effectively understand biomolecules using natural language and design them according to human intentions.
This presents a gap between AI's computational capacity and researchers' needs to apply it to real-world problems. 
Consider the scenario of a researcher tasked with designing a new drug to target a protein involved in a complex disease. Traditionally, this process involves navigating vast amounts of biomolecular data, interpreting biological and chemical relationships, and iterating through trial-and-error to engineer molecules with specific properties. While AI has enhanced many aspects of this workflow, current tools often struggle to align molecular complexity with intuitive, human-driven goals articulated in natural language.}

{To fully unlock the potential of AI in biomolecular science, there is an urgent need for systems that can seamlessly bridge the gap between biomolecular data and human intention. Such a system would allow researchers to articulate their design goals in natural language and receive molecular outputs that meet precise biological needs—whether it be generating a drug molecule tailored to a specific protein or designing an enzyme optimized for a chemical process. This alignment of AI with human expertise and intuition is crucial for advancing research in areas where creativity and expert knowledge play a central role.
}

Notably, Large Language Models (LLMs)~\cite{DBLP:journals/corr/abs-2302-09419} have demonstrated their impressive capacity to understand human intention and generate human-like responses helpfully and safely~\cite{DBLP:journals/corr/abs-2307-09288,DBLP:journals/corr/abs-2303-08774}. This capability derives from its vast number of learnable parameters, training on expansive corpora, and strong alignment with human preferences.
Nevertheless, despite their potential to comprehend and design biomolecules in natural language following human intentions, leveraging LLMs for biomolecular research is still in its nascent stage and presents several challenges~\cite{zhang2025scientific}.
Firstly, biomolecular research demands substantial specialized knowledge. General-purpose LLMs may not possess deep insights into this field because they are not tailored for specialized domains. Consequently,
the lexical and semantic gap between the natural language and the language used to describe biomolecules presents its challenges. For instance, ``C'' could represent an alphabet letter in English, a carbon atom in a chemical molecule, or cysteine in the context of amino acids in proteins. Such semantic discrepancies might confuse models that are not specifically designed for these contexts.
Secondly, unlike the primarily textual focus of general LLMs, biomolecules are intrinsically multimodal. For example, molecules are often represented by sequences like SMILES~\cite{weininger1988smiles} or SELFIES~\cite{DBLP:journals/mlst/KrennHNFA20}. Also, molecules can be inherently depicted as 2D graphs, featuring atoms as nodes and chemical bonds as edges, or as 3D structures, noting the spatial arrangements of atoms. Similarly, proteins are described through FASTA~\cite{pearson1994using} sequences for their amino acid composition, and they also have 3D structures, essential for understanding their interactions and functions in space~\cite{jumper2021highly}. 
The effective utilization of this multimodal data presents a unique challenge for text-focused LLMs.
{Lastly, and most importantly, general LLMs struggle to align human intention in biomolecular tasks.}
To achieve the desired performance in a specific domain, LLMs need to be trained with alignment to acquire task-specific knowledge and patterns~\cite{DBLP:conf/nips/Ouyang0JAWMZASR22}. Instruction plays a critical role in this alignment process. By providing carefully curated, domain-specific instructions, LLMs are guided to develop a deeper understanding and more precise execution of specific tasks.
In particular, to follow instructions for biomolecules, mastering the alignment between natural language, molecules, and proteins
is essential. Complex tasks such as designing molecules for target proteins or creating enzymes for substrates necessitate simultaneous processing of natural language, molecules, and proteins.

Although several recent endeavors~\cite{DBLP:conf/emnlp/EdwardsLRHCJ22,DBLP:conf/acl/WangZDQZLC24,DBLP:conf/emnlp/PeiZZWGWXY23,DBLP:conf/iclr/FangL0LH0FC24,DBLP:conf/acl/PeiWGLFZ00024,DBLP:journals/corr/abs-2308-09442,DBLP:conf/iclr/LiuWYW0GX24} have sought to tailor the LLMs for biomolecular tasks via extensive instructions, they encounter two primary limitations: (1) They typically align natural language with either molecules or proteins, but not both, lacking {any-to-any} alignment. (2) They also fall short in processing multimodal biomolecules, failing to align the multimodal data with natural language.

\begin{figure*}[!t]
\centering
\includegraphics[width=\linewidth]{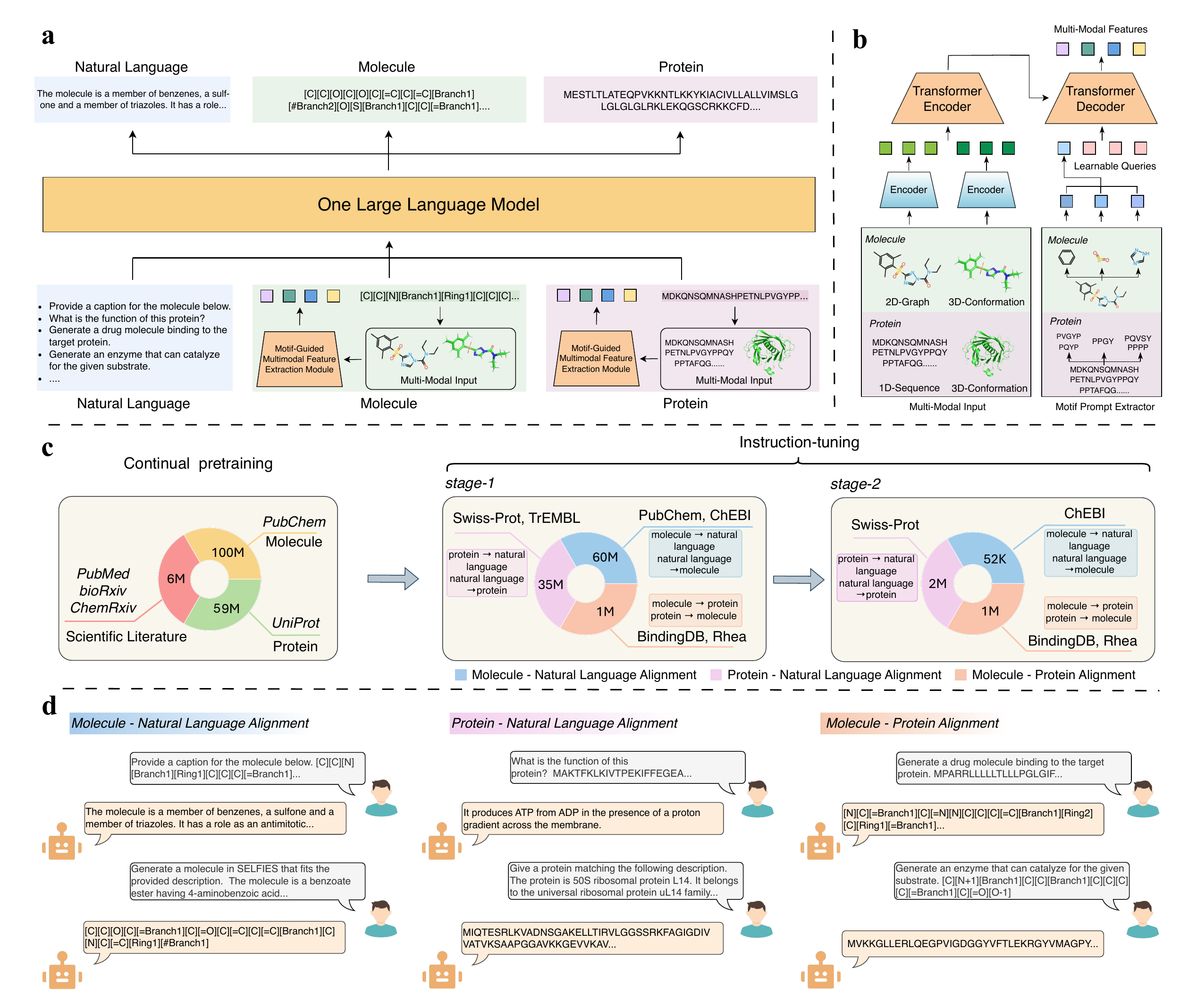}
\caption{\textbf{An overview of \modelname{}.} \textbf{a}, \modelname{} is a unified multimodal Large Language Model for natural language, molecules, and proteins. {It can accept inputs in the form of natural language text, multimodal molecule and protein data and generate outputs as natural language, molecules, or proteins in the textual form.} \textbf{b}, The Motif-Guided Multimodal Feature Extraction Module processes 2D graphs and 3D structures of molecules, as well as 1D sequences and 3D structures of proteins. Pre-trained modality-specific encoders obtain representations from these inputs, which are then processed by a Transformer Encoder. The Transformer Decoder, using motif prompts and learnable queries, produces multimodal features for integration into the language model.
The 2D molecule illustration is derived using SmilesDrawer~\cite{probst2018smilesdrawer}, and the 3D molecule and protein illustrations are from PyMOL~\cite{PyMOL}.
\textbf{c}, We collect datasets on a hundred-million scale, categorized into continual pretraining data and instruction-tuning data. Instruction-tuning is used to achieve an any-to-any 
alignment between molecule, protein, and natural language. A two-stage instruction-tuning paradigm enables the model to learn from low-quality extensive data {(stage-1)} to high-quality refined data {(stage-2)}. \textbf{d}, \modelname{} achieves alignment between molecules and natural language, proteins and natural language, as well as molecules and proteins. This enables it to follow human instructions, facilitating the understanding and design of biomolecules. The icons are from Flaticon.com. } 
\label{fig:framework}
\end{figure*}

In this study, we introduce \modelname{}, which exhibits the following key characteristics:
\begin{itemize}
    \item \textbf{Biomolecular instruction following}. \modelname{} integrates natural language and biomolecules within one Large Language Model, {becoming the first to} achieve {any-to-any} pairwise alignment between natural language, molecules, and proteins.
    By leveraging a curated hundred-million scale instruction-tuning dataset, the model is empowered to understand and design biomolecules according to human intention.
    \item \textbf{Multimodal data understanding}. We propose a motif-guided multimodal feature extraction module. It utilizes pre-trained encoders to capture various features, including 2D topological and 3D geometric details of molecules, and 1D sequence and 3D geometric properties of proteins. Also, we design a motif prompt extractor, which leverages biological knowledge embedded in motifs to guide the multimodal feature fusion.
    \item {\textbf{Serving as a research copilot and supporting practical biomolecular tasks 
    }}. {The value of \modelname{} is its role as a digital research assistant, supporting researchers in biomolecular studies and discoveries. It excels in employing natural language processing to explore biomolecules, such as answering questions related to protein functions or designing novel molecules based on textual descriptions. Moreover, \modelname{} demonstrates its potential in solving practical tasks, such as drug discovery and enzyme design.}
\end{itemize}

{In the experiments, 
\modelname{} is thoroughly assessed for its proficiency in understanding and designing molecules and proteins following human instructions,
showing its ability to align natural language with biomolecules. {Specifically, \modelname{} achieves an overall improvement of 12\% in understanding and designing molecule and protein benchmark tasks.} Additionally, we explore the model's application in generating drug molecules aimed at specific target proteins and designing enzyme proteins for particular substrates. 
Experimental results reveal that drug molecules designed by \modelname{} exhibit an improvement of 10\% in binding affinity, while the enzymes it designs achieve an ESP Score~\cite{kroll2023general} of 70.4, making it the only method to surpass the enzyme-substrate interaction threshold of 60.0 recommended by the ESP developer.
These exercises confirm \modelname{}'s applicability in practical biomolecular research scenarios.
Overall, these results not only validate the broad generalization of \modelname{} as a Large Language Model that bridges natural language and biomolecules, but also underscore its potential for a wide range of applications within the life sciences.}

\section*{Results}

\subsection*{Overview of \modelname{}}
The overview of \modelname{} is presented in Figure~\ref{fig:framework}a. {\modelname{} is a unified multimodal Large Language Model that simultaneously handles natural language, molecules, and proteins. It accepts inputs in natural language or multimodal molecules and proteins, and generates natural language, molecules, or proteins in textual form.}
To process multimodal data of both molecules and proteins, we develop a Motif-Guided Multimodal Feature Extraction Module (Figure~\ref{fig:framework}b). This module employs a Transformer Encoder-Decoder structure~\cite{DBLP:conf/nips/VaswaniSPUJGKP17}.
It encodes 2D graphs and 3D structures for molecules and 1D sequences along with 3D structures for proteins.
Using pre-trained lightweight encoders, the multimodal inputs are encoded into corresponding representations, and subsequently fed into a Transformer Encoder. {In the Transformer Decoder, we extract the biological knowledge in the motifs using a motif prompt extractor, which serves as guiding information for the fusion of multimodal features. The fused features are then integrated into the language model.}
For {model} training, outlined in Figure~\ref{fig:framework}c, a wide range of data is collected, comprising a continual pretraining dataset and an instruction-tuning dataset. The continual pretraining dataset comprises molecules, proteins and natural language texts derived from scientific literature. The instruction-tuning dataset consists of data pairs between natural language, molecules, and proteins.
The training process is divided into two stages. First, continual pretraining is employed to augment domain-specific knowledge in the field of biomolecular scientific research.
Then, instruction-tuning is performed to achieve an {any-to-any} pairwise alignment among natural language, molecules, and proteins. {We employ a staged instruction-tuning pipeline, learning from large-scale data (stage-1) to refined data (stage-2) to gradually improve performance.}
As a result, \modelname{} aligns natural language, molecules, and proteins in an any-to-any manner, demonstrating competency across a broad spectrum of biomolecular tasks, as shown in Figure~\ref{fig:framework}d. This includes solving practical challenges, like the discovery of molecule drugs for target proteins, and the design of enzymes for specific substrates, {following human intention}.

\subsection*{{\modelname{} can understand and design molecules}}
\paragraph{Experimental Setup.}

We evaluate \modelname{}'s capability in understanding and designing molecules through two tasks: (1) molecule captioning, which involves generating a textual description for a molecule; and (2) description-based molecule generation, where a molecule is generated based on a provided textual description. These tasks are introduced by ref.~\citenum{DBLP:conf/emnlp/EdwardsLRHCJ22}, and use the ChEBI dataset~\cite{hastings2016chebi}, which contains molecules and their corresponding descriptions including structure, function, origin, etc.
{The evaluation metrics are consistent with those in ref.~\citenum{DBLP:conf/emnlp/EdwardsLRHCJ22} and described in Methods.}
In this experiment, we evaluate two types of baselines. 
{(1) Generalist language models.  We assess two variants of general-purpose models. First, GPT-3.5 (zero-shot), a commonly used Large Language Model. Second, GPT-3.5 (10-shot MolReGPT) and GPT-4 (10-shot MolReGPT), which are adaptations of GPT-3.5 and GPT-4, respectively, using MolReGPT's few-shot in-context learning approach~\cite{molregpt}. (2) Molecule-specific enhanced language models. These models are further fine-tuned for molecular tasks, building on generalist models. Baselines include ChemDFM~\cite{DBLP:journals/corr/abs-2401-14818}, InstructMol~\cite{DBLP:journals/corr/abs-2311-16208}, MolT5~\cite{DBLP:conf/emnlp/EdwardsLRHCJ22}, BioT5~\cite{DBLP:conf/emnlp/PeiZZWGWXY23}, and BioT5+~\cite{DBLP:conf/acl/PeiWGLFZ00024}.

\paragraph{Results.}
Quantitative results on molecule captioning and molecule generation are in Table~\ref{tab:molecule-caption} and Table~\ref{tab:molecule-generation}, respectively. 
Molecule-specific models outperform generalist language models due to finetuning on domain-specific instruction datasets, which align natural language with chemical knowledge. This highlights that general models lack domain expertise, which can be addressed through instruction-tuning.
Notably, the experimental results demonstrate that \modelname{} performs best across almost all evaluation metrics. Specifically, for the molecule captioning task, \modelname{} yields an average improvement of 0.9\% across all metrics. In description-based molecule generation task, the exact match accuracy (EXACT) of generated molecules increases by 0.7\%. Furthermore, an average improvement of 2.0\% is observed in molecular fingerprint similarity metrics (MACCS FTS, RDK FTS, and MORGAN FTS). These results indicate that \modelname{} exhibits higher accuracy and efficacy in both understanding and generating chemical molecular information. We attribute this success to the extensive use of high-quality instruction data, which enables the model to comprehensively align molecules and natural language and achieve superior performance across molecular tasks.

\begin{table}[t]
\centering
\caption{Performance comparison on molecule captioning task. (\textuparrow) / (\textdownarrow) denotes that a higher / lower value is better. The best performance is marked in bold.}
\scalebox{0.75}{{
\begin{tabular}{lcccccc}
\toprule
                           & BLEU-2 (\textuparrow) & BLEU-4 (\textuparrow) & ROUGE-1 (\textuparrow) & ROUGE-2 (\textuparrow) & ROUGE-L (\textuparrow) & METEOR (\textuparrow) \\ \midrule
GPT-3.5 (zero-shot) & 10.3 & 5.0 & 26.1 & 8.8 & 20.4 & 16.1 \\
GPT-3.5 (10-shot MolReGPT) & 56.5 & 48.2 & 62.3 & 45.0 & 54.3 & 58.5 \\
GPT-4 (10-shot MolReGPT) & 60.7 & 52.5 & 63.4 & 47.6 & 56.2 & 61.0 \\ \midrule
ChemDFM & 32.1 & 26.5 & 49.0 & 37.4 & 48.3 & 40.2 \\
InstructMol & 47.5 & 37.1 & 56.6 & 39.4 & 50.2 & 50.9 \\
MolT5 & 64.4 & 57.2 & 70.8 & 58.4 & 65.3 & 68.1 \\
BioT5 & 63.5 & 55.6 & 69.2 & 55.9 & 63.3 & 65.6 \\
BioT5+ & \textbf{66.6} & 59.1 & 71.0 & 58.4 & 65.0 & 68.1 \\
\modelname{} & 66.3 & \textbf{59.3} & \textbf{72.0} & \textbf{60.1} & \textbf{66.8} & \textbf{69.1} \\ \bottomrule
\end{tabular}
}}
\label{tab:molecule-caption}
\end{table}

\begin{table}[t]
\centering
\caption{Performance comparison on description-based molecule generation task. (\textuparrow) / (\textdownarrow) denotes that a higher / lower value is better. The best performance is marked in bold.}
\scalebox{0.65}{{
\begin{tabular}{lcccccccc}
\toprule
                                 & BLEU (\textuparrow)  & EXACT (\textuparrow) & LEVENSHTEIN (\textdownarrow) & MACCS FTS (\textuparrow) & RDK FTS (\textuparrow) & MORGAN FTS (\textuparrow) & FCD (\textdownarrow)  & VALIDITY (\textuparrow) \\ \midrule
GPT-3.5 (zero-shot) & 48.9 & 1.9 & 52.13 & 70.5 & 46.2 & 36.7 & 2.05 & 80.2 \\
GPT-3.5 (10-shot MolReGPT) & 79.0 & 13.9 & 24.91 & 84.7 & 70.8 & 62.4 & 0.57 & 88.7 \\
GPT-4 (10-shot MolReGPT) & 85.7 & 28.0 & 17.14 & 90.3 & 80.5 & 73.9 & 0.41 & 89.9 \\ \midrule
ChemDFM & 83.9 & 43.2 & 16.90 & 90.1 & 82.9 & 75.9 & - & 97.6\\
MolT5 & 85.4 & 31.1 & 16.07 & 83.4 & 74.6 & 68.4 & 1.20 & 90.5 \\
BioT5 & 86.7 & 41.3 & 15.10 & 88.6 & 80.1 & 73.4 & 0.43 & \textbf{100.0} \\
BioT5+ & 87.2 & 52.2 & \textbf{12.78} & 90.7 & 83.5 & 77.9 & 0.35 & \textbf{100.0} \\
InstructBioMol & \textbf{87.7} & \textbf{52.9} & 13.65 & \textbf{91.8} & \textbf{85.8} & \textbf{80.5} & \textbf{0.24} & 99.0 \\
\bottomrule
\end{tabular}
}}
\label{tab:molecule-generation}
\end{table}

\begin{figure*}[!t]
\centering
\includegraphics[width=\linewidth]{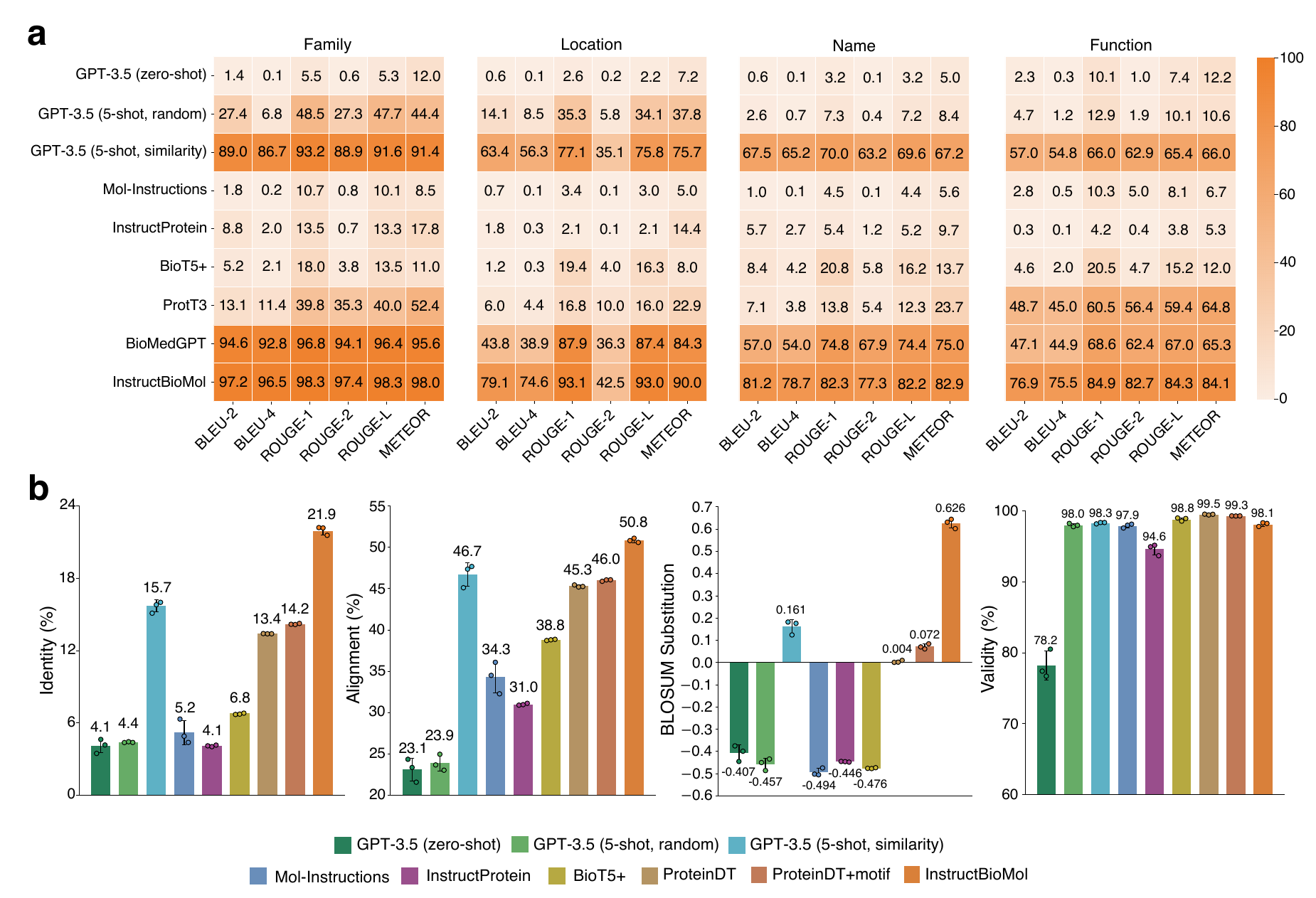}
\caption{\textbf{Model performance on protein understanding and design benchmarks.} \textbf{a}, Regarding protein understanding, models are evaluated by examining their performance in answering questions related to protein family, subcellular location, name, and function. The evaluation metrics employed include BLEU-2, BLEU-4, ROUGE-1, ROUGE-2, ROUGE-L, and METEOR. \textbf{b}, For description-based protein generation task, the accuracy and biological validity of the generated proteins are assessed using metrics including Identity, Alignment, BLOSUM Substitution, and Validity. The number of experimental replicates is $n$=3, and the error bars are standard deviation.}
\label{fig:result-prottext}
\end{figure*}

\begin{figure*}[!t]
\centering
\includegraphics[width=\linewidth]{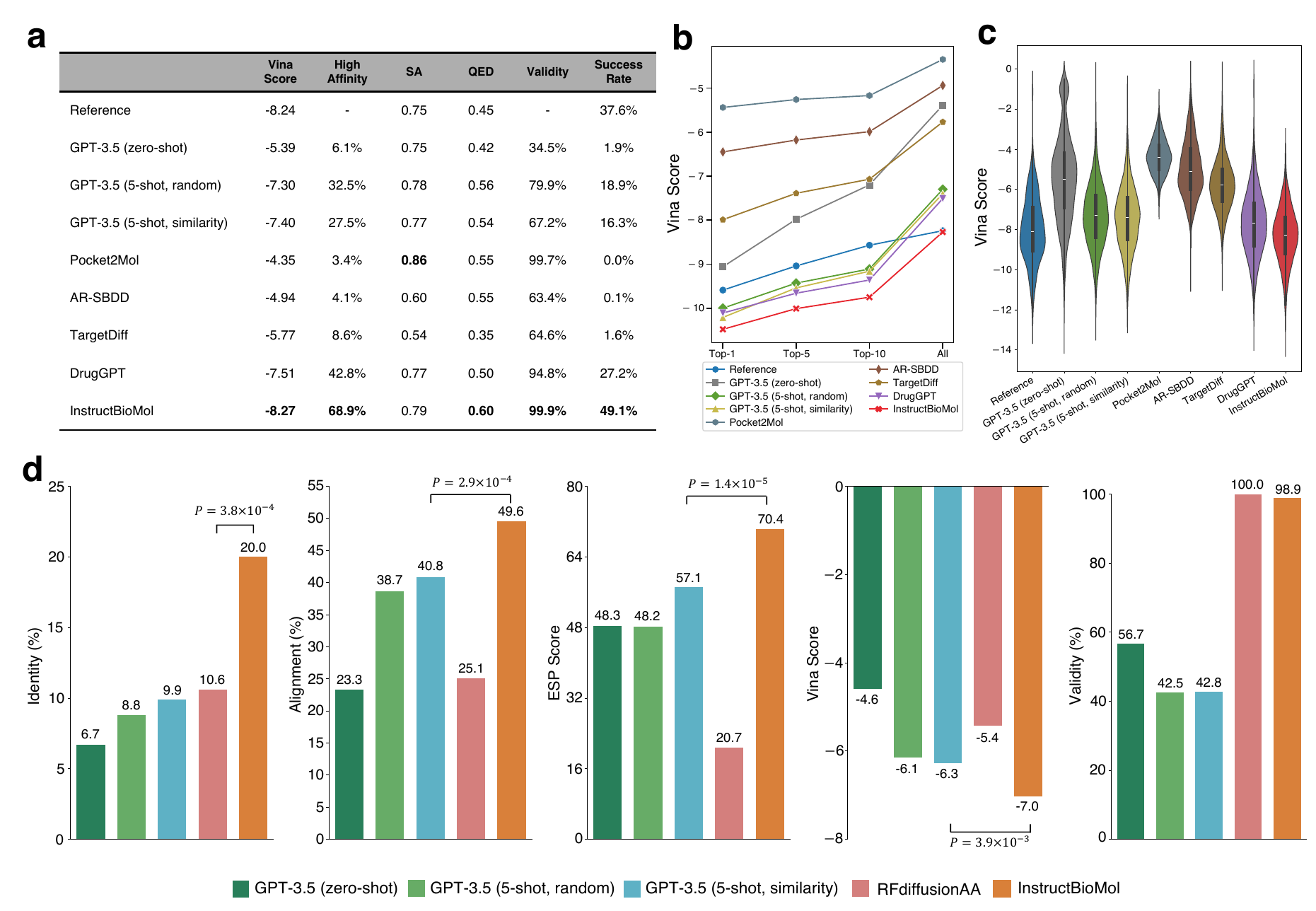}
\caption{\textbf{Model performance on drug discovery and enzyme design.} \textbf{a-c}, Performance comparison on drug discovery. \textbf{(a)}, The generated drug-like molecules are evaluated from multiple perspectives, including binding affinity (e.g., Vina Score and High Affinity), general properties (SA, QED, and Validity), and an overall evaluation metric (Success Rate). \textbf{b-c}, A detailed analysis of Vina Scores is presented, including (\textbf{b}) top-1, top-5, top-10, and all Vina Scores, as well as (\textbf{c}) the distribution of Vina Scores for all generated molecules. ($n$=10,000, median as centre line, 25th to 75th percentiles as bounds of the box, whiskers extending to 1.5 times the interquartile range from the bounds of the box.)
{\textbf{d}, Performance comparison with statistical significance on enzyme design. Evaluation metrics include similarity metrics Identity and Alignment; interaction metrics ESP Score and Vina Score; and Validity.}
Statistical significance is determined by two-sided independent sample t-test.}
\label{fig:result-molprot}
\end{figure*}

\subsection*{{\modelname{} can understand and design proteins}}
\paragraph{Experimental Setup.}
We evaluate the model's ability to understand and design proteins in the following two tasks: (1) answering questions about the properties of proteins, including protein family, subcellular location, official name, and function; (2) generating protein sequences based on the textual descriptions.
{The evaluation metrics are described in Methods.}
In the experiments, we compare the following baseline methods: (1) the general-purpose language model GPT-3.5. GPT-3.5 is evaluated in three variants: GPT-3.5 (zero-shot), GPT-3.5 (5-shot, random), and GPT-3.5 (5-shot, similarity). The GPT-3.5 (5-shot, random) and GPT-3.5 (5-shot, similarity) variants utilize an in-context learning paradigm (detailed in Supplementary Information Section 2), where the former randomly selects 5 examples from the training set as prompts, and the latter selects the 5 most similar examples to the given query. (2) Protein-specific enhanced language models, which extend general language models to protein-related tasks, including Mol-Instructions~\cite{DBLP:conf/iclr/FangL0LH0FC24}, InstructProtein~\cite{DBLP:conf/acl/WangZDQZLC24}, BioT5+~\cite{DBLP:conf/acl/PeiWGLFZ00024}, ProtT3~\cite{DBLP:conf/acl/LiuZ0ZWKC24}, BioMedGPT~\cite{DBLP:journals/corr/abs-2308-09442}, ProteinDT~\cite{DBLP:journals/corr/abs-2302-04611}, {and ProteinDT+motif, a variant of ProteinDT that incorporates motif embeddings during training}.

\paragraph{Results.}
Quantitative results on answering protein properties and description-based protein generation are in Figure~\ref{fig:result-prottext}a and Figure~\ref{fig:result-prottext}b, respectively.
Based on the experimental results, we have reached the following conclusions: {Firstly, \modelname{} demonstrates the best performance in both tasks. In tasks related to answering questions about protein properties, \modelname{} outperforms previous state-of-the-art (SOTA) methods by 13.5\% on average. For protein generation tasks, \modelname{} achieves 6.2\%, 4.1\%, and 0.465 improvements in Identity, Alignment, and BLOSUM Substitution, respectively, compared to previous SOTAs, and comparable validity of the generated proteins.} 
Secondly, domain-specific instruction alignment proves highly effective. In protein property question tasks, {\modelname{} outperforms GPT-3.5 (zero-shot) significantly, and in protein generation, it shows a 17.8\% increase in Identity and a 27.7\% improvement in Alignment.} These results highlight that models tailored with domain-specific instructions significantly outperform general models in specialized tasks, particularly in fields like protein engineering, where integrating domain knowledge greatly enhances both practicality and accuracy.
%

\subsection*{{\modelname{} enables target protein-based drug discovery}}
\paragraph{Experimental Setup.}
Designing molecules that can bind to specific proteins is one of the most challenging tasks in drug discovery~\cite{anderson2003process}. The chemical space is vast, yet the subset of molecules with desirable biological activity is relatively small. Drug discovery typically involves searching this expansive space for molecules that can bind to specific targets, such as disease-related proteins. As a research copilot, \modelname{} can design molecule drugs for target proteins following human intention, thereby reducing the time and cost of drug development. {In this experiment, evaluation metrics are detailed in Methods.}
To make a comparison, we take three variants of GPT-3.5 (zero-shot, 5-shot random, and 5-shot similarity), {Pocket2Mol~\cite{DBLP:conf/icml/PengLGXPM22}, AR-SBDD~\cite{DBLP:conf/nips/LuoGMP21}, TargetDiff~\cite{DBLP:conf/iclr/GuanQPS0M23}} and DrugGPT~\cite{li2023druggpt} as baselines, and also take molecules in the test set as the Reference baseline. 100 target proteins are selected in the test set. For each target protein, 100 molecules are generated.
\paragraph{Results.}
Figure~\ref{fig:result-molprot}a presents the experimental results. where \modelname{} demonstrates superior performance across three key dimensions: binding affinity, general properties, and overall assessment. Specifically, it improves High Affinity and Success Rate by 25.9\% and 21.9\%, respectively, compared to previous state-of-the-art (SOTA) methods.  This highlights \modelname{}'s enhanced capability in generating drug-like molecules with high affinity for target proteins and favorable intrinsic properties. Additionally, it achieves an outstanding generation validity of 99.9\%. Figure~\ref{fig:result-molprot}b illustrates the average Vina Scores for top-1, top-5, top-10, and all generated molecules.
{It is evident that \modelname{} consistently outperforms other methods in these settings, achieving a $P$-value of 0.05 for outperforming the second-best model (DrugGPT) in the top-1 Vina Score. Moreover, it is the only approach where the average scores for all generated molecules exceed the reference values.}
This suggests that the quality of the molecules generated by \modelname{} is comparable to the ground truth in the dataset.
{Moreover, \modelname{} proves to be the most effective method for designing molecules with the best Vina Scores for most target proteins. As shown in Supplementary Figure 1a, \modelname{} achieves the best performance on 35\% of the targets.}
Figure~\ref{fig:result-molprot}c shows the distribution of Vina Scores for all generated molecules. The distribution reveals that molecules generated by \modelname{} have a lower mean and reduced variance, further confirming that the overall quality of these molecules is superior to that of other methods.


\subsection*{{\modelname{} enables target substrate-based enzyme design}}
\paragraph{Experimental Setup.}
Enzymes, as biological catalysts, can accelerate chemical reactions in various biological processes~\cite{bar2011moderately}. In enzymatic reactions, substrates are the small molecules that are catalytically converted by enzymes. By binding to specific substrates and acting upon them, enzymes significantly enhance the conversion efficiency of the substrates. Designing enzymes that can bind to specific substrates is a crucial and challenging research problem. \modelname{} can assist researchers in designing protein enzymes for specific substrates, thereby advancing the progress of efficient enzyme design. To establish experimental validation, we split a test set containing 100 substrates, and for each target substrate, 100 enzyme proteins are generated.
{The evaluation metrics employed in this experiment are detailed in Methods.}
For comparison, we employ three variants of GPT-3.5 (zero-shot, 5-shot random, 5-shot similarity) and RFdiffusionAA~\cite{krishna2024generalized} as baselines. 

\paragraph{Results.}
The performance of the generated protein enzymes across various evaluation metrics is presented in Figure~\ref{fig:result-molprot}d. \modelname{} demonstrates the best performance in terms of similarity to ground truth, interaction capability with substrates, and exhibits superior generation validity. {Specifically, \modelname{} achieves improvements of 13.3\% in ESP Score~\cite{kroll2023general} and 0.7 in Vina Score (with $P$-values of $1.4\times 10^{-5}$ and $3.9 \times 10^{-3}$), indicating a stronger potential for substrate binding compared to baseline methods.} {Notably, \modelname{} attains an ESP Score of 70.4, making it the only method to surpass the enzyme-substrate interaction threshold of 60.0 recommended by the ESP developer. This demonstrates that enzymes designed by \modelname{} can bind their corresponding substrates with high affinity.}
{Supplementary Figure 1b} further analyzes the top-1 ESP Score of the proteins generated for each substrate, revealing that \modelname{} achieves the best performance on {66\%} of the substrates. Additionally, {Supplementary Figure 1c} presents the top-1 Vina Score on each substrate, with  \modelname{} attaining the best performance on {89\%} of the substrates.
These findings suggest that \modelname{} holds significant potential in generating highly efficient and specific protein enzymes, offering more effective solutions for fields such as biocatalysis. 

\begin{figure*}[!t]
\centering
\includegraphics[width=\linewidth]{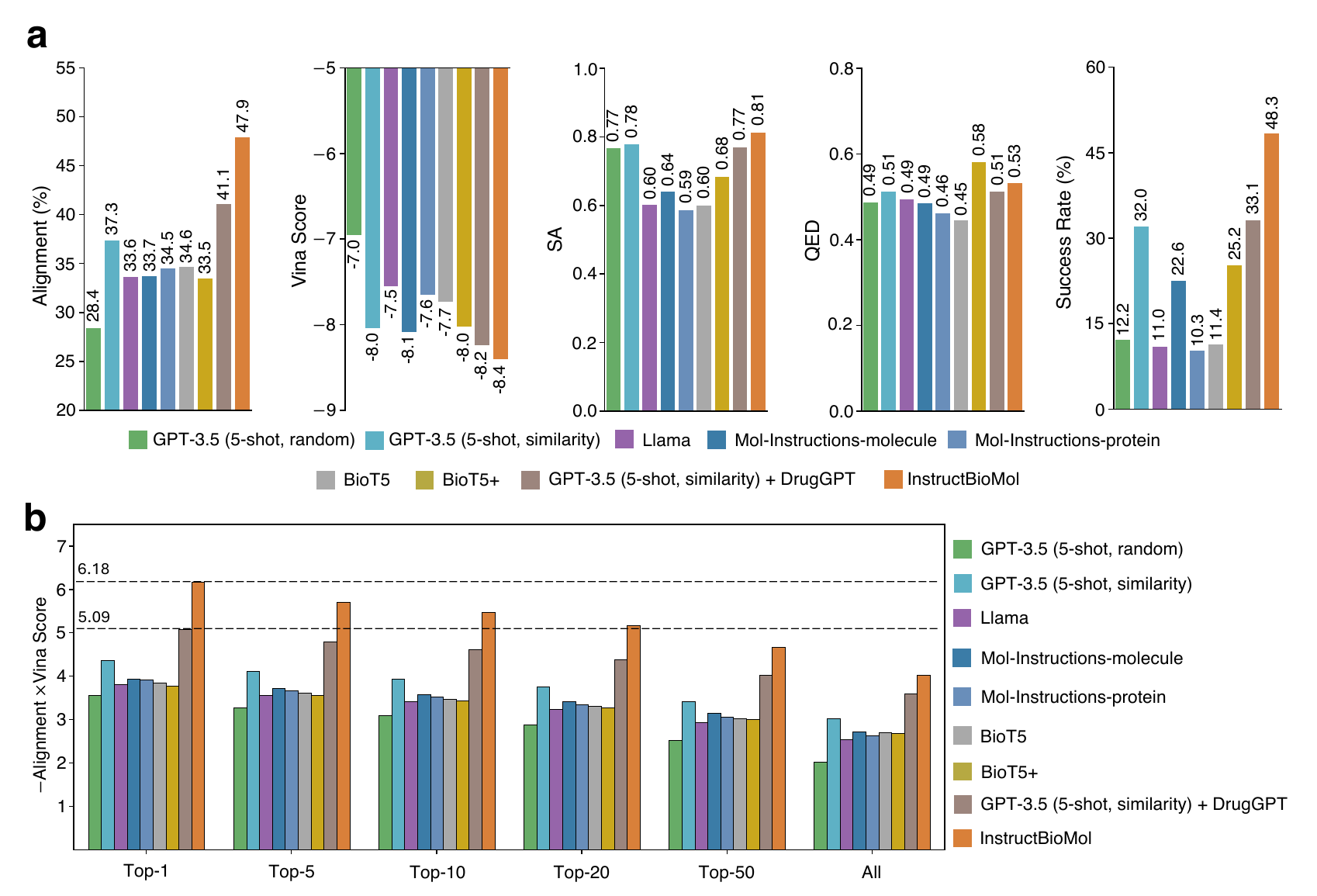}
\caption{\textbf{Performance on description-based protein-molecule pairs generation.} \textbf{a}, The generated proteins and molecules are evaluated from three perspectives: protein assessment (Alignment), molecule assessment (Vina Score, SA, QED), and overall assessment (Success Rate). \textbf{b}, $-$ Alignment × Vina Score across top-1, top-5, top-10, top-20, top-50, and all generated protein-molecule pairs for each description. A higher value indicates better quality of generated protein-molecule pairs.}
\label{fig:t2mp}
\end{figure*}

\subsection*{{\modelname{} excels at designing protein and molecule simultaneously from descriptions}}
\paragraph{{Experimental Setup.}}
{As a unified Large Language Model, \modelname{} has demonstrated exceptional alignment between natural language, molecules, and proteins. We further explore the advantages of this unified model by generating proteins and binding molecules simultaneously from textual descriptions. We split a training set and a test set, and the test set includes 100 target protein descriptions. In evaluation, for each description, 100 protein-molecule pairs are generated. The evaluation metrics are detailed in Methods. We adopt LoRA~\cite{DBLP:conf/iclr/HuSWALWWC22} fine-tuning on \modelname{} for this task. Baselines consist of three groups. (1) GPT-3.5, which includes two variants: GPT-3.5 (5-shot, random) and GPT-3.5 (5-shot, similarity). (2) Fine-tuned Large Language Models. This includes Llama~\cite{DBLP:journals/corr/abs-2307-09288}, Mol-Instructions-molecule, Mol-Instructions-protein, BioT5, and BioT5+. All of these models are fine-tuned on the training set. (3) Sequential ensemble approach: GPT-3.5 (5-shot, similarity) + DrugGPT, which combines the most competitive techniques for generating proteins from natural language and generating molecules from proteins. First, GPT-3.5 (5-shot, similarity) is used to generate proteins from descriptions, followed by the use of DrugGPT to generate molecules based on the proteins.}


\paragraph{{Results.}}
{We present the experimental results in Figure~\ref{fig:t2mp}, which demonstrate that \modelname{} significantly outperforms the baseline methods. In protein generation, Alignment improves by 6.8\%; in molecule generation, both Vina Score and SA achieve the best performance. Overall, the Success Rate reaches 48.3\%, which is 15.2\% higher than the baseline. Additionally, we analyze the top-n ($-$ Alignment $\times$ Vina Score) for each description, where a higher value indicates better quality of both generated proteins and molecules. We find that even in the top-20 results, our method still slightly outperforms the top-1 score of the strongest baseline, further proving the efficiency of \modelname{}. We also observe that baselines based on fine-tuned Large Language Models do not perform as well. While Mol-Instructions-molecule, Mol-Instructions-protein, and \modelname{} are all trained on Llama, their performances vary significantly. Specifically, Mol-Instructions-molecule, which is trained solely on text-molecule data, excels in molecule-related metrics like Vina Score, SA, and QED, but struggles with protein generation. In contrast, Mol-Instructions-protein, trained only on text-protein data, demonstrates strong protein generation capabilities with a high Alignment but performs poorly on molecule-related metrics. By comparison, \modelname{} shows a marked performance advantage. This performance discrepancy clearly highlights that models specialized in either molecules or proteins struggle to meet the complex demands of tasks that require simultaneous integration of natural language, molecular, and protein data. Benefiting from achieving any-to-any alignment across natural language, molecules, and proteins, \modelname{} can learn across different data types more effectively, capturing cross-modal semantic relationships.}

\section*{Discussion}
{In this study, we propose \modelname{}, a multimodal Large Language Model capable of following human instructions for understanding and designing biomolecules. To address the limitation that general-purpose language models cannot handle multimodal biomolecular data, we design a motif-guided multimodal feature extraction module. This module extracts multimodal features from biomolecules and leverages the knowledge embedded in motifs to guide the fusion of these features, which are then integrated into the language model. During training, we employ a training paradigm that involves ``continual pretraining followed by instruction-tuning'', based on extensive pretraining and instruction-tuning data. In instruction-tuning, we adopt a staged strategy to progressively reduce the data size while enhancing data quality.  Through comprehensive instruction-tuning, \modelname{} becomes the first model capable of achieving any-to-any alignment between natural language, molecules, and proteins. Our experiments demonstrate the effectiveness of these alignments across a range of tasks involving natural language, molecules, and proteins. \modelname{} is not only capable of understanding and designing molecules or proteins following human intention, but it can also design drug-like molecules for target proteins or enzyme catalysts for reaction substrates. This indicates \modelname's potential as a research copilot, offering valuable insights and inspiration to researchers, with practical applications in drug and enzyme design.
}

{One limitation of \modelname{} lies in the constraints imposed by computational resources, preventing it from fully supporting all biomolecules, such as DNA and RNA. Furthermore, it has not been comprehensively trained across all biomolecular tasks, which limits its ability to handle certain additional tasks, such as chemical reaction prediction. However, based on the current model architecture and training framework, \modelname{} exhibits strong extensibility. By incorporating more multimodal encoders and expanding its vocabulary, it can enhance its encoding and generation capabilities for other biomolecules, and it can be easily adapted to new tasks through additional instruction data. Another concern is the profound implications and potential risks associated with the integration of Large Language Models and biomolecules. Ensuring alignment between LLMs and human ethics is crucial. For instance, when utilizing LLMs to design novel biomolecules, adherence to strict ethical guidelines is essential to avoid irresponsible experimentation and potential biosafety hazards. In the future, we plan to enhance the alignment of \modelname{} with human values and ethics, ensuring its consistency with societal norms and enabling it to inspire biomolecular innovations safely and effectively.}

{We believe the core value of \modelname{} lies in pioneering a new paradigm for processing biomolecular data using Large Language Models, showcasing the potential of general intelligence in handling diverse tasks in one model. With the increase in computational resources, the enrichment of training data, and the enhanced alignment with human ethics, \modelname{} is expected to evolve and support a broader range of tasks effectively and safely, laying the groundwork for advancing Artificial General Intelligence (AGI) in scientific research.}


\renewcommand{\tablename}{}
\renewcommand{\thetable}{Extended Data Table \arabic{table}}
\setcounter{table}{0}

\section*{Methods}

\subsection*{Model Architecture}
{The architecture of \modelname{} (Figure~\ref{fig:framework}a) consists of two components: the Motif-Guided Multimodal Feature Extraction Module (Figure~\ref{fig:framework}b) and the {Biomolecular Vocabulary-expanded Language Model}. The former is designed to extract multimodal features of biomolecules, while the latter handles a unified processing of textual natural language, molecule and protein data, as well as the extracted multimodal features. Specifically, in the Motif-Guided Multimodal Feature Extraction Module, we employ lightweight frozen pre-trained encoders to extract features from each modality separately, and leverage the biological knowledge embedded in motifs to guide the fusion of these multimodal features. Within the {Biomolecular Vocabulary-expanded Language Model}, to mitigate potential interference among data from different domains, we expand the vocabulary to accommodate molecules and proteins, and standardize the input format for their multimodal features.}
\subsubsection*{Motif-Guided Multimodal Feature Extraction Module}
Biomolecules exhibit inherent multimodality, characterized by diverse sequential and structural representations across various domains~\cite{DBLP:conf/icml/GilmerSRVD17,DBLP:conf/iclr/ZhouGDZXWZK23,lin2022language,jumper2021highly}. This complexity cannot be fully captured by any single modality in isolation. In this module (Figure~\ref{fig:framework}b), we incorporate 2D-graph and 3D-structure for molecules, alongside 1D-sequence and 3D-structure for proteins to leverage the multitude of perspectives available.
2D-graph of molecules highlights the basic skeleton, while 3D-structure provides insights into molecular docking and interaction. For proteins, 1D-sequence delineates the fundamental arrangement of amino acids, and 3D-structure unlocks understanding of functional sites and foldings.
We utilize frozen pre-trained encoders to process each modality separately, and then leverage the inherent biological knowledge within motifs to guide multimodal feature fusion, which enhances comprehension and processing of complex biological data.
\paragraph{Multimodal Inputs and Encoders.}  
For molecules, the 2D-graph modality is defined as $m_{2D}=(V, E)$, where $V$ stands for atomic nodes and $E$ represents chemical bonds between these atoms. The 3D-structure modality is defined as $m_{3D}=(V,\mathbf{C})$, with $V$ indicating a set of atoms and $\mathbf{C} \in \mathbb{R}^{|V| \times 3}$ representing the spatial coordinates of these atoms.
We leverage a pre-trained 5-layer GIN~\cite{DBLP:conf/iclr/XuHLJ19, DBLP:conf/iclr/HuLGZLPL20} as the 2D-graph encoder $f_m^{2D}$, and Geoformer~\cite{DBLP:conf/nips/WangLWS0L23} as the 3D-structure encoder $f_m^{3D}$, to derive the respective modality inputs' representations:
\begin{equation}
    \mathbf{H}^{2D}_m=f_m^{2D}(m_{2D}), \quad \mathbf{H}^{3D}_m=f_m^{3D}(m_{3D}),
    \label{eq:mol-encoder}
\end{equation}
where $\mathbf{H}^{2D}_m \in \mathbb{R}^{|V| \times d_m^{2D}}$ and $\mathbf{H}^{3D}_m \in \mathbb{R}^{|V| \times d_m^{3D}}$ are the obtained molecular 2D and 3D representations, respectively. 
For proteins, the 1D-sequence modality is characterized by $p_{1D}=(s_1, s_2, ..., s_N)$, where each $s_i$ is an amino acid.
And the 3D-structure modality can be represented as $p_{3D}=(S, \mathbf{C})$, where $S$ is the amino acid sequence and $\mathbf{C} \in \mathbb{R}^{N \times 4 \times 3}$ denotes the coordinates of four backbone atoms (N, C, CA, O) in each amino acid.
Here, we adopt pre-trained ESM2-35M~\cite{lin2022language} and SaProt-35M~\cite{su2024saprot} as the 1D-sequence encoder $f_p^{1D}$ and 3D-structure encoder $f_p^{3D}$, respectively, to encode the two modalities of proteins:
\begin{equation}
    \mathbf{H}^{1D}_p=f_p^{1D}(p_{1D}), \quad \mathbf{H}^{3D}_p=f_p^{3D}(p_{3D}),
    \label{eq:prot-encoder}
\end{equation}
where $\mathbf{H}^{1D}_p \in \mathbb{R}^{N \times d_p^{1D}}$ and $\mathbf{H}^{3D}_p \in \mathbb{R}^{N \times d_p^{3D}}$ represent the obtained protein 1D and 3D representations, respectively.

\paragraph{Motif Prompt Extractor.}

In molecules, a motif often represents a functional group or substructure, playing a crucial role in determining molecular function and structure~\cite{DBLP:conf/nips/ZhangLWLL21,li2023knowledge}. Similarly, in proteins, motifs are sequences of consecutive amino acids carrying specific biological functions, forming foundational elements for protein functionality~\cite{grant2011fimo}.
To integrate the essential prior knowledge within motifs, we introduce a motif prompt. The motif prompt is designed to highlight key regions within biomolecules that are pivotal for understanding their function and interaction. By acting as a conditional input, it guides the multimodal feature extraction process towards features that are relevant to the identified motifs, thereby increasing the biological relevance of the extracted features.
Specifically, we denote the motifs in a molecule as $\mathbf{T}_m=[t_1^m, t_2^m, ..., t_{N_m}^m]$, and motifs in a protein as $\mathbf{T}_p=[t_1^p, t_2^p, ..., t_{N_p}^p]$, where $t_i \in \{0,1\}$ indicates the presence ($t_i$=1) or absence ($t_i$=0) of the $i$-th motif in the molecule or protein, and $N_m$ and $N_p$ represent the total counts of predefined motifs in molecules and proteins, respectively.
Subsequently, the motif prompt for molecules or proteins is computed as:
\begin{equation}
    \mathbf{P}_{m}=\mathbf{T}_m\mathbf{M}_m, \quad \mathbf{P}_{p}=\mathbf{T}_p\mathbf{M}_p,
    \label{eq:prompt}
\end{equation}
where $\mathbf{P}_m \in \mathbb{R}^{d}$ and $\mathbf{P}_p \in \mathbb{R}^{d}$ are motif prompt of molecule and protein, respectively. $\mathbf{M}_m \in \mathbb{R}^{N_m \times d}$ and $\mathbf{M}_p \in \mathbb{R}^{N_p \times d}$ are two learnable matrices. In detail, we obtain $\mathbf{T}_m$ by computing the Functional-Class FingerPrint (FCFP)~\cite{rogers2010extended} of a molecule with a radius of 2 and a length of 1024. The motifs of protein are collected from the UniProt~\cite{boeckmann2003swiss} database, and details are described in Supplementary Information Section 1.

\paragraph{Joint Multimodal Feature Extraction.}
A Transformer~\cite{DBLP:conf/nips/VaswaniSPUJGKP17} Encoder-Decoder architecture is employed for effective extraction and fusion of features from biomolecular various modalities. The reason for the choice of the Transformer architecture is that the inherent self-attention mechanism allows for the dynamic weighting of various parts of the input data, enabling to focus on the most relevant features across and within modalities. The flexibility of the Encoder-Decoder facilitates the cross-modal integration of features, with the encoder capturing the salient features and the decoder synergizing those to construct a fused multimodal representation.
Firstly, the single-model representations obtained from Eq.~\ref{eq:mol-encoder} and Eq.~\ref{eq:prot-encoder} are transformed and then concatenated:
\begin{equation}
\begin{split}
    &{\mathbf{H}^{2D}_{m}}^{\prime}=\mathrm{MLP}(\mathrm{LayerNorm}(\mathbf{H}^{2D}_{m})), \quad 
    {\mathbf{H}^{3D}_{m}}^{\prime}=\mathrm{MLP}(\mathrm{LayerNorm}(\mathbf{H}^{3D}_{m})), \quad \mathbf{H}_m=[{\mathbf{H}^{2D}_{m}}^{\prime} \oplus {\mathbf{H}^{3D}_{m}}^{\prime}], \\
    &{\mathbf{H}^{1D}_{p}}^{\prime}=\mathrm{MLP}(\mathrm{LayerNorm}(\mathbf{H}^{1D}_{p})), \quad 
    {\mathbf{H}^{3D}_{p}}^{\prime}=\mathrm{MLP}(\mathrm{LayerNorm}(\mathbf{H}^{3D}_{p})), \quad \mathbf{H}_p=[{\mathbf{H}^{1D}_{p}}^{\prime} \oplus {\mathbf{H}^{3D}_{p}}^{\prime}],
    \label{eq:enc-input}
\end{split}
\end{equation}
where $\mathbf{H}_m \in \mathbb{R}^{2|V| \times d}$ and $\mathbf{H}_p \in \mathbb{R}^{2N \times d}$ are used as the inputs of the Transformer Encoder for molecule and protein, respectively, and $\oplus$ denotes the concatenation operation.
We utilize the motif prompt obtained from Eq.~\ref{eq:prompt} as the initial input to the Transformer Decoder.
By directing the focus of the Transformer-Decoder toward these motifs, this approach endeavors to anchor the multimodal feature extraction in biologically significant referents. This ensures that the resultant fused features are not only data-derived but also deeply rooted in the biological realities of molecular and protein functionalities.
Additionally, the input to the Transformer Decoder includes a sequence of learnable queries. Formally, the joint multimodal feature extraction is defined as:
\begin{equation}
    \mathbf{Z}=[\mathbf{z}_1, \mathbf{z}_2, ..., \mathbf{z}_{(1+N_q)}]=\mathrm{Transformer}(\mathrm{Enc}(\mathbf{H}),\mathrm{Dec}([\mathbf{P}\oplus \mathbf{Q}])).
    \label{eq:multimodal}
\end{equation}
To simplify, subscripts are omitted since molecules and proteins undergo the same processing. Here, $\mathbf{P}$ and $\mathbf{H}$ are derived from Eq.~\ref{eq:prompt} and Eq.~\ref{eq:enc-input} respectively. $\mathbf{Q}$  denotes a sequence of learnable queries as $\mathbf{Q} = [\mathbf{q}_1, \mathbf{q}_2, ..., \mathbf{q}_{N_q}]  \in \mathbb{R}^{N_q \times d}$, and $\mathbf{Z} \in \mathbb{R}^{(1+N_q) \times d}$ is the extracted multimodal features.

\subsubsection*{{Biomolecular Vocabulary-expanded Language Model}}
\paragraph{Language Model Backbone.} \modelname{} is designed to be compatible with any GPT-style~\cite{radford2018improving} language model. In this study, we specifically adopt {Llama-2-7B~\cite{DBLP:journals/corr/abs-2307-09288}} for further training.

\paragraph{Expanding Vocabulary.} In this work, we use SELFIES~\cite{DBLP:journals/mlst/KrennHNFA20} to represent molecules and FASTA~\cite{pearson1994using} (sequence of amino acids) to represent proteins. Despite their utility, a notable conflict arises among natural language, molecules, and proteins, where identical tokens may imply entirely different meanings.
For example, the token \textrm{"C"} in English simply refers to the letter C, but in molecular contexts, it represents a carbon atom, and in protein sequences, it denotes cysteine. This ambiguity prevents natural language vocabularies from distinguishing these entities effectively.
Hence, we introduce extended vocabulary for molecules and proteins, integrating them with the original natural language vocabulary.
Specifically, for molecules, we utilize the pair of brackets within SELFIES along with the meaningful group of atoms they encapsulate as a token. For instance, \textrm{"[C]"} denotes a Carbon atom. For proteins, we introduce a specific prefix "<p>" for each amino acid, such as "<p>C" for cysteine. 
Furthermore, we introduce specialized tokens to differentiate between modalities. These include \textrm{"<SELFIES>"}, \textrm{"</SELFIES>"} for molecule SELFIES sequence,  \textrm{"<FASTA>"}, \textrm{"</FASTA>"} for protein amino acid sequence, as well as \textrm{"<MOL>"}, \textrm{"</MOL>"} \textrm{"<PROT>"} \textrm{"</PROT>"} signify outputs from the Motif-Guided Multimodal Feature Extraction Module for molecules and proteins, respectively. This deliberate separation of different modalities ensures the preservation of each modality's intrinsic integrity and prevents model confusion regarding the meanings of different modalities.

\paragraph{Input Formation.}
{
For molecules and proteins, we concatenate multimodal features $\mathbf{Z}=[\mathbf{z}_1,\mathbf{z}_2,...,\mathbf{z}_{(N_q+1)}]$ obtained from Eq.~\ref{eq:multimodal} with sequence-modality input, and label them with special tokens:}

\begin{equation}
\begin{aligned}
    \label{eq:x_biomol}
    x_{multimodal\_mol} &= \textrm{<MOL>} \;[\mathbf{z}_1,\mathbf{z}_2,...,\mathbf{z}_{(N_q+1)}] \; \textrm{</MOL> \; <SELFIES> \; [SELFIES Sequence] \;</SELFIES>}, \\
    x_{multimodal\_prot} &= \textrm{<PROT>} \;[\mathbf{z}_1,\mathbf{z}_2,...,\mathbf{z}_{(N_q+1)}] \; \textrm{</PROT> \;<FASTA>\;[FASTA Sequence]\;</FASTA>}.
\end{aligned}
\end{equation}
{
$x_{multimodal\_mol}$ and $x_{multimodal\_prot}$ denote molecules and proteins, respectively. In practice, we treat $\mathbf{Z}$ as an additional feature, concatenated with the text representation derived from tokenizing the input and passing it through the LLM embedding layer. 
The inputs also could include natural language instructions  $x_{insuction}$, e.g., "\textit{What is the function of this protein}", and description text $x_{text}$, e.g., "\textit{The molecule is a member of benzenes, a sulfone and a member of triazoles.}". The specific composition of inputs depends on the task.
For instance, generating molecular descriptions involves $x_{instruction}$ and $x_{multimodal}$, while generating molecules from descriptions uses $x_{instruction}$ and $x_{text}$.
The detailed task-specific input forms are provided in \ref{tab:instruction-tuning-dataset}.}

\subsection*{Data Collection}
We collect datasets on a hundred-million-scale, including a continual pretraining dataset (\ref{tab:self-supervised-traing-dataset}), and an instruction-tuning dataset (\ref{tab:instruction-tuning-dataset}).
The continual pretraining dataset comprises molecules and proteins in textual format, and natural language texts derived from scientific literature, enabling the model to develop a foundational adapting to biomolecular research.
The instruction-tuning dataset contains various alignment pairs: molecule-natural language, protein-natural language, and molecule-protein (Figure~\ref{fig:framework}d), achieving any-to-any alignment among molecules, proteins and natural language. Additionally, both molecular and protein data in the instruction-tuning dataset are multimodal, incorporating 2-D and 3-D structures of molecules and 3-D structures of proteins.
%
%

\subsubsection*{{Continual} Pretraining Dataset}

For molecules, we collect 115 million entries from PubChem~\cite{kim2016pubchem}, filtering out those with atomic numbers exceeding 50, resulting in 100 million entries and 4 billion tokens after tokenization. For proteins, we use the Uniref50 dataset~\cite{suzek2007uniref}, which comprises 59 million entries and 15 billion tokens after tokenization.
Recognizing the limitations of general language models trained on generalized corpora lacking biomolecular insights, we augment our data collection with literature specific to the research domain. 
Abstracts of scientific papers are collected across several sources, including PubMed~\cite{white2020pubmed}, bioRxiv~\cite{sever2019biorxiv}, and ChemRxiv~\cite{mudrak2022five}, to enrich our training corpus. The incorporation of literature from these repositories enhances the domain-specific knowledge of \modelname{}. This subset includes 6 million abstracts, resulting in 8 billion tokens.

\subsubsection*{Instruction-tuning Dataset}
\paragraph{Molecule-Natural Language Pairs.}
The dataset is sourced from two databases.
The first is from PubChem, where we collect molecules with the IUPAC (International Union of Pure and Applied Chemistry)~\cite{mcnaught1997compendium} name. This naming convention establishes a standardized nomenclature, fostering uniformity and clarity across the chemical community. 30 million molecules with IUPAC names are sampled from the filtered set in the previous step.
The second source is ChEBI~\cite{hastings2016chebi} data from ref.~\citenum{DBLP:conf/emnlp/EdwardsLRHCJ22}, comprising molecules alongside their descriptions. These descriptions encapsulate various facets of molecular structure, function, synthesis methodologies, etc.

\paragraph{Protein-Natural Language Pairs.}
Data originate from two databases: SwissProt and TrEMBL~\cite{boeckmann2003swiss}.
These databases provide textual descriptions of proteins, covering four key aspects: name, family, location, and function. 
We utilize SwissProt data collected in ref.~\citenum{DBLP:journals/corr/abs-2308-09442} and curate TrEMBL data from the UniProt~\cite{uniprot2018uniprot} database.
To ensure data quality and diversity, we filter TrEMBL dataset using UniRef50, resulting in approximately 25 million proteins. 
Additionally, for proteins described in at least three of four aspects, we consolidate these descriptions into a comprehensive summary using ChatGPT, creating data aligned from natural language to proteins.

\paragraph{Molecule-Protein Pairs.}
For molecule and protein pairs, we focus on two key applications: the discovery of drug-like molecules for specific target proteins and the design of enzyme proteins to catalyze specific substrates. Specifically, for generating molecules to specific target proteins, we use data from BindingDB~\cite{gilson2016bindingdb} collected in ref.~\citenum{uludougan2022exploiting}. BindingDB is a public database primarily focusing on the interactions between proteins, identified as potential targets, and small, drug-like molecular ligands.
On the other hand, for designing an enzyme to catalyze a particular substrate, we draw upon data from the Rhea~\cite{bansal2022rhea} database collected in ref.~\citenum{kroll2023general}. Rhea is an expert-curated database of chemical reactions of biological interest, where enzyme-catalyzed reactions are curated from peer-reviewed literature.

\paragraph{Multimodal Data.}
The extraction of multimodal features necessitates access to diverse data types relating to molecules and proteins. To accomplish this, we employ RDKit~\cite{landrum2013rdkit} to convert molecules to 2D-graph and optimize them using ETKDG~\cite{riniker2015better} and Merck Molecular Force Field~\cite{halgren1996merck} to obtain 3D-structure. For 3D-structure of proteins, we download the predicted 3D structures from the AlphaFold Protein Structure Database~\cite{varadi2022alphafold} via the UniProt ID of each protein. 
This guarantees that the molecule and protein data have rich multimodal characteristics, laying a solid foundation for their application in downstream processes.

\subsection*{Training Strategy}
We start with a pretrained language model and continue pretraining it on the continual pretraining dataset in a {self-supervised causal language modeling objective~\cite{radford2018improving}}. Subsequently, we employ instruction-tuning, to establish an any-to-any alignment among natural language, molecules, and proteins.
This involves aligning specific instructions and inputs with appropriate responses, represented as $(x_{instruction},x_{input}) \rightarrow y$, where $x_{input}$ may include multimodal molecules $x_{multimodal\_mol}$, proteins $x_{multimodal\_prot}$ defined in Eq.~\ref{eq:x_biomol},
or natural language in textual format $x_{text}$, and $y$ denotes corresponding responses such as natural language $y_{text}$, molecular sequences $y_{mol}$, or protein sequences $y_{protein}$. To achieve a thorough alignment, we introduce a bidirectional alignment task for each pairwise alignment among natural language, molecules and proteins.
For example, for molecule-natural language pairs, one task generates textual descriptions from molecular data:  $(x_{instruction}, x_{multimodal\_mol}) \rightarrow y_{text}$, and another task generates molecules from descriptions: $(x_{instruction}, x_{text}) \rightarrow y_{mol}$. The instruction-tuning is optimized under a causal language modeling objective:
\begin{equation}
    \min_{\theta}\mathcal{L}_{CE}\left( \mathrm{LM}\left( x_{instruction},  x_{input} \right), y\right),
\end{equation}
where $\mathcal{L}_{CE}$ is cross-entropy loss, $\theta$ is all the model parameters except four frozen pre-trained multimodal encoders, $\mathrm{LM}(\cdot)$ denotes the language model's prediction, and $y$ is the label.

\paragraph{Two-Stage Instruction-tuning.}
Despite the collection of a broad range of data, the quality of this data exhibits considerable variability across sources.
For example, the ChEBI database offers a broader spectrum of molecular descriptions compared to the natural-language-like structure descriptions provided by IUPAC names in PubChem. Similarly, while data within the SwissProt undergo meticulous manual curation, entries in the TrEMBL do not benefit from such rigorous calibration.
The tradeoff between the scale and quality of data poses significant challenges to model performance and generalizability.
To address this issue, we adopt a two-stage instruction-tuning strategy designed to exploit the extensive data initially, then progressively direct the focus towards the insights offered by higher-quality datasets. 
Initially, in stage-1, the model is trained across all the available instructions. This stage leverages the diversity and volume of data to build a foundation on biomolecular alignment. 
Subsequently, in stage-2, the model undergoes further fine-tuning on a subset of higher-quality data.
This approach harnesses both the expansive coverage of lower-quality data and the precision inherent in high-quality data, facilitating an efficient and effective utilization of the dataset. The scale and specific details of the different datasets used in the two stages are presented in Figure~\ref{fig:framework}c and \ref{tab:instruction-tuning-dataset}.


\subsection*{Implementation Details}
{We use Pytorch~\cite{DBLP:conf/nips/PaszkeGMLBCKLGA19} to implement the model.} {The model is trained on 8 80G NVIDIA H800 GPUs.} Additionally, we adopt the DeepSpeed ZeRO-1~\cite{DBLP:conf/sc/RajbhandariRRH20} and BF16 for computational efficiency.
The total number of training steps is 1.5 million. The steps for continual pretraining, stage-1 instruction-tuning and stage-2 instruction-tuning are 600,000, 500,000, and 400,000 respectively. We use the AdamW optimizer with $(\beta_1, \beta_2)$ set to $(0.9,0.95)$. We follow a linear learning rate schedule, warming up from 0 to maximum learning rate 1e-5 over the first 2,000 steps, and decaying the final learning rate down to 0. During training, all parameters are trainable except for the modality encoders $f_m^{2D}$, $f_m^{3D}$, $f_p^{1D}$, $f_p^{3D}$ in Equations~\ref{eq:mol-encoder} and \ref{eq:prot-encoder}, with the total trainable parameters being 6.8B. The ratio of different datasets sampled during training is controlled using hyper-parameters, {which are detailed in Supplementary Table 1.}

\subsection*{Evaluation Tasks and Datasets}
To validate the model's capability in molecule understanding and design, two tasks are performed: molecule captioning and description-based molecule generation. The dataset we use is consistent with ref.~\citenum{DBLP:conf/emnlp/EdwardsLRHCJ22}. To assess the model's ability to understand and design proteins, we conduct experiments on protein property answering and description-based protein generation tasks. The data splits used in the experiments follow ref.~\citenum{DBLP:journals/corr/abs-2308-09442} , with 3,000 samples selected as the test set for each task. Specifically, the protein properties include family, subcellular location, name, and function. For proteins containing at least three of these properties, we use ChatGPT to combine them into a complete description, which serves as the dataset for protein design based on the description. For the task of designing molecule drugs targeting proteins, we select 100 proteins as the test set. Similarly, for the task of designing enzymes based on substrates, we select 100 enzymes as the test set. Details of datasets are provided in the Supplementary Information Section 3.

\subsection*{Evaluation Metrics}
\subsubsection*{Evaluation Metrics for Molecule Captioning}
We leverage standard natural language generation metrics such as BLEU~\cite{DBLP:conf/acl/PapineniRWZ02}, ROUGE~\cite{lin2004rouge}, and METEOR~\cite{DBLP:conf/acl/BanerjeeL05} to evaluate molecule captioning following ref.~\citenum{DBLP:conf/emnlp/EdwardsLRHCJ22}. These metrics measure how closely the generated captions match the reference captions.
\subsubsection*{Evaluation Metrics for Description-based Molecule Generation}
The following several types of metrics proposed in ref.~\citenum{DBLP:conf/emnlp/EdwardsLRHCJ22} are used to evaluate the task of generating molecules from textual descriptions:
\paragraph{BLEU.} Similar to BLEU scores in natural language processing, the SMILES BLEU score measures the overlap between the generated molecules and the reference in SMILES strings.
\paragraph{EXACT.} This metric checks for exact matches between the generated and reference molecules. It provides a strict measure of accuracy.
\paragraph{LEVENSHTEIN.} Levenshtein distance~\cite{miller2009levenshtein} measures the number of single-character edits required to transform the generated molecules in SMILES format into the reference string. A lower Levenshtein distance indicates a closer match.
\paragraph{Fingerprint Metrics.} We use three types of fingerprint metrics—MACCS FTS, RDK FTS, and Morgan FTS. These use the MACCS fingerprint~\cite{durant2002reoptimization}, RDK fingerprint~\cite{schneider2015get}, and Morgan fingerprint~\cite{rogers2010extended}, respectively. And then calculating the Tanimoto similarity~\cite{bajusz2015tanimoto} between the fingerprints of the generated and reference molecules, providing a measure of how structurally similar the generated molecules are to the reference molecules.
\paragraph{FCD.} FCD (Fréchet ChemNet Distance)~\cite{preuer2018frechet} compares the distributions of features derived from ChemNet between the generated and reference molecules. A lower value indicates a closer match.
\paragraph{Validity.} This metric assesses the percentage of generated molecules that are syntactically valid according to chemical rules.

\subsubsection*{Evaluation Metrics for Protein Property Answering}
Considering that both protein property answering and molecule captioning are natural language generation tasks, we adopt the evaluation metrics of the molecule captioning task to assess this task following ref.~\citenum{DBLP:conf/emnlp/EdwardsLRHCJ22,DBLP:journals/corr/abs-2308-09442}.
\subsubsection*{Evaluation Metrics for Description-based Protein Generation}

\paragraph{Identity.} This metric is designed to measure the similarity between two protein sequences by calculating their percentage identity. It first counts the number of identical residues by comparing each corresponding residue in the reference protein $p_{ref}$ and the generated protein $p_{gen}$. Then it normalizes the number of identical residues by the sum of the lengths of both sequences. Formally, 
\begin{equation}
    Identity = \frac{2\times identical\_residues}{len(p_{ref})+len(p_{gen})} \times 100.
\end{equation}
This formula yields a normalized value that ranges between 0 and 100,  where a value of 100 indicates perfect identity, and a value of 0 indicates no identity.
\paragraph{Alignment.} This metric assesses the similarity between two protein sequences by leveraging the alignment scoring, which performs sequence alignment using the Smith-Waterman algorithm~\cite{smith1981identification} to identify regions of alignment subsequences between the reference protein $p_{ref}$ and the generated protein $p_{gen}$. The alignment focuses on finding the highest-scoring subsequences, which allows the comparison of potentially functionally or structurally significant regions. This metric is computed based on the alignment score, and then normalized by the combined lengths of both sequences:
\begin{equation}
    Alignment = \frac{2\times alignment\_score}{len(p_{ref})+len(p_{gen})} \times 100.  
\end{equation}
This normalization accounts for the length variations of the proteins and ensures the metric ranges between 0 and 100, with 100 indicating perfect alignment similarity and 0 indicating no alignment.

\paragraph{BLOSUM Substitution.} This metric calculates the similarity between the reference protein $p_{ref}$ and the generated protein $p_{gen}$ using a BLOSUM45~\cite{henikoff1992amino} substitution matrix-based scoring approach. This substitution matrix is commonly employed to assess the evolutionary similarity of protein sequences by providing scores for each possible pair of amino acids based on observed substitution frequencies in homologous proteins. For each pair of residues at a corresponding position, we first retrieve the substitution score from the BLOSUM45 matrix. Then, the total score, representing the cumulative similarity of all residue pairs, is normalized by the length of the two sequences:
\begin{equation}
    BLOSUM\_Substitution=\frac{2\times \sum substitution\_matrix(a,b)}{len(p_{ref})+len(p_{gen})},
\end{equation}
where $a$ and $b$ are amino acids from $p_{ref}$ and $p_{gen}$, respectively, and $substitution\_matrix(a,b)$ denotes the substitution score for the pair. When $substitution\_matrix(a,b) > 0$, it indicates that the substitution of one amino acid for another occurs more frequently in related proteins than would be expected by chance, suggesting conservative substitutions and likely preserving protein structure and function. On the other hand, when $substitution\_matrix(a,b) < 0$, it signifies that the substitution is less common, implying a disruptive effect on protein function or structure.

\paragraph{Validity.} This metric is employed to evaluate the valid proportion of the generated proteins, assessing whether the generated proteins are composed of amino acid sequences.

\subsubsection*{Evaluation Metrics for Target Protein-based Drug Discovery}
For drug discovery, we evaluate the generated molecules from three perspectives following ref.~\citenum{DBLP:conf/iclr/GuanQPS0M23,DBLP:conf/icml/QuQSG0Z0M24}:
\paragraph{Target Binding Affinity.} 
Binding affinity reflects the interaction strength between the generated molecules and the target protein. \textbf{Vina Score} is used to estimate this affinity, with lower scores indicating stronger binding. Specifically, we first retrieve protein structures from AlphaFold Protein Structure Database~\cite{varadi2022alphafold}, then use DiffDock-L~\cite{corso2024discovery} to estimate the protein-molecule complex structures. Qvina~\cite{alhossary2015fast} is then employed to compute the scores. Additionally, we introduce the \textbf{High Affinity} metric to measure the proportion of generated molecules that achieve better binding scores than reference molecules within the test set.
\paragraph{Molecular Property.}
General molecular properties, such as \textbf{QED} (Quantitative Estimation of Drug-likeness)~\cite{bickerton2012quantifying} and \textbf{SA}~(Synthetic Accessibility)~\cite{ertl2009estimation}, are utilized to evaluate the drug-likeness and synthetic accessibility of molecules. QED provides an assessment of a molecule's potential as a drug by considering parameters like molecular weight, lipophilicity, and polar surface area, with scores ranging from 0 to 1; higher scores indicate greater drug-likeness. SA quantifies the ease of molecule synthesis, also on a scale from 0 to 1, with higher scores reflecting simpler synthetic processes. Furthermore, \textbf{Validity} is employed to determine the proportion of generated molecules that are syntactically valid according to chemical rules.

\paragraph{Overall Assessment.}
Following ref.~\citenum{DBLP:conf/icml/QuQSG0Z0M24}, we use the \textbf{Success Rate} to assess the quality of the generated molecules by considering multiple factors, including binding affinity, drug-likeness, and synthetic accessibility. A molecule is successful if it meets specific thresholds for Vina Score, QED, and SA (Vina Score $<$ -8.18, QED $>$ 0.25, SA $>0.59$).

\subsubsection*{Evaluation Metrics for Target Substrate-based Enzyme Design}
In the task of enzyme design, since generations are protein sequences, we choose to use \textbf{Identity} and \textbf{Alignment} metrics to assess the similarity between the generated and reference proteins. To assess the interaction between enzyme proteins and substrates, we employ two metrics: \textbf{Vina Score} and \textbf{ESP Score}~\cite{kroll2023general}. Vina Score is used to quantify the strength of the interaction between the designed enzyme and its substrate. A lower value indicates a stronger interaction. Specifically, when calculating the Vina Score, we use DiffDock-L to obtain the complex structure and then use Qvina to obtain the corresponding score. ESP Score is another metric for evaluating enzyme-substrate interactions. This score ranges from 0 to 100, with higher scores indicating stronger interactions. We evaluate the model's optimal performance by calculating the average top-1 Identity, Alignment, Vina score, and ESP score for all substrate-specific designed proteins. Additionally, we use the \textbf{Validity} metric to evaluate the biological validity of the generated proteins.

\subsubsection*{{Evaluation Metrics for Designing Protein and Molecule from Descriptions}}

{The evaluation metrics are organized into three groups: (1) \textbf{Protein evaluation}: This focuses on calculating the \textbf{Alignment} between the generated proteins and the ground truth. (2) \textbf{Molecule evaluation}: This group evaluates the generated molecules using the Vina Score to assess the binding affinity between the molecules and proteins, as well as assesses molecular properties including \textbf{QED} and \textbf{SA}. (3) \textbf{Overall evaluation}: This group combines both generated proteins and molecules. Specifically, we use the \textbf{Success Rate} to assess the proportion of successful generations, considering factors such as Alignment, Vina Score, QED, and SA. A generation is considered successful if it meets the following specific thresholds: Alignment $>$ 30\%, Vina Score $>$ -8.18, QED $>$ 0.25, and SA $>$ 0.59.}

\section*{Data availability}
Dataset used in this study is available in Zenodo at \href{https://doi.org/10.5281/zenodo.15303508}{https://doi.org/10.5281/zenodo.15303508} (ref.~\citenum{zhuang_2025_15303508}).

\section*{Code availability}
The source code of this study is available in GitHub at \href{https://github.com/HICAI-ZJU/InstructBioMol}{https://github.com/HICAI-ZJU/InstructBioMol}~(ref.~\citenum{xiang_zhuang_2025_15335654}).

\section*{Acknowledgements}
This work is funded by NSFCU23B2055 (H.C.), NSFC2302433 (Q.Z.), NSFCU23A20496 (Q.Z.), the Fundamental Research Funds for the Central Universities (226-2023-00138, H.C.), Zhejiang Provincial “Jianbing” “Lingyan” Research and Development Program of China (2025C01097, K.D. and Q.Z.), Zhejiang Provincial Natural Science Foundation of China (LQ24F020007, Q.Z.) and Hangzhou West Lake Pearl Project Leading Innovative Youth Team Project (TD2023017, K.D.).

\section*{Author Contributions Statement}
X.Z., K.D., Q.Z., and H.C. conceived the study. X.Z. developed the method, implemented the code, and conducted the experiments. T.L. participated in benchmarking some baseline models. X.Z., Y.J., X.L., and Z.X. contributed to dataset collection. K.D., Z.W., M.Q., K.F., J.W., Q.Z., and H.C. provided critical suggestions on methodology and experiments.
All authors wrote the paper, reviewed it, and approved the final paper.
\section*{Competing Interests Statement}
The authors declare no competing interests.

\clearpage
\addtocontents{toc}{\protect\setcounter{tocdepth}{-10}}
\bibliography{main}
\addtocontents{toc}{\protect\setcounter{tocdepth}{1}}
\clearpage

\begin{table}[h]
\centering
\caption{Statistics of {continual} pretraining dataset.}
\begin{tabular}{lccc}
\toprule
Data type        & Entries & Tokens  & Source                    \\ \midrule
molecule         & 100M & 4B & PubChem                   \\ \midrule
protein          & 59M & 15B & Uniref50                  \\ \midrule
natural language & 6M & 8B  & PubMed, bioRxiv, ChemRxiv \\ \bottomrule
\end{tabular}

\label{tab:self-supervised-traing-dataset}
\end{table}

\begin{sidewaystable}
[]
\caption{Statistics of instruction-tuning dataset.}
\begin{tabular}{llllcc}
\toprule
Sub-dataset            & Task type                       & Scale  & Instruction                                                             & stage-1 & stage-2 \\ \midrule
\multicolumn{6}{l}{\textit{\textbf{Molecule - Natural Language Alignment}}}                                                                                                                      \\ \midrule
PubChem                & $(x_{instruction},x_{multimodal\_mol})\rightarrow y_{text}$ & 30M  & Give the IUPAC name of the following molecule.                      &   \checkmark     &        \\ \midrule
PubChem                & $(x_{instruction},x_{text})\rightarrow y_{mol}$ & 30M  & Generate a molecule in SELFIES that fits the provided IUPAC name.   &   \checkmark     &        \\ \midrule
ChEBI                  & $(x_{instruction},x_{multimodal\_mol})\rightarrow y_{text}$ & 26K  & Provide a caption for the molecule below.                           &    \checkmark    &    \checkmark    \\ \midrule
ChEBI                  & $(x_{instruction},x_{text})\rightarrow y_{mol}$ & 26K  & Generate a molecule in SELFIES that fits the provided description.  &    \checkmark    &     \checkmark   \\ \midrule
\multicolumn{6}{l}{\textit{\textbf{Protein - Natural Language Alignment}}}                                                                                                                       \\ \midrule
TrEMBL\_Name           & $(x_{instruction},x_{multimodal\_protein})\rightarrow y_{text}$  & 25M  & What is the official name of this protein?                          &    \checkmark    &        \\ \midrule
TrEMBL\_Family         & $(x_{instruction},x_{multimodal\_protein})\rightarrow y_{text}$  & 5M   & What is the protein family that this protein belongs to?            &   \checkmark     &        \\ \midrule
TrEMBL\_Locaction      & $(x_{instruction},x_{multimodal\_protein})\rightarrow y_{text}$  & 2M   & What is the subcellular location of this protein?                   &     \checkmark   &        \\ \midrule
TrEMBL\_Function       & $(x_{instruction},x_{multimodal\_protein})\rightarrow y_{text}$  & 1M   & What is the function of this protein?                                 &    \checkmark    &        \\ \midrule
TrEMBL\_Description    & $(x_{instruction},x_{text})\rightarrow y_{prot}$  & 2M   & Generate a protein matching the following description.               &     \checkmark   &        \\ \midrule
SwissProt\_Name        & $(x_{instruction},x_{multimodal\_protein})\rightarrow y_{text}$  & 455K & What is the official name of this protein?                          &   \checkmark     &     \checkmark   \\ \midrule
SwissProt\_Family      & $(x_{instruction},x_{multimodal\_protein})\rightarrow y_{text}$  & 370K & What is the protein family that this protein belongs to?            &    \checkmark    &     \checkmark   \\ \midrule
SwissProt\_Location    & $(x_{instruction},x_{multimodal\_protein})\rightarrow y_{text}$  & 275K & What is the subcellular location of this protein?                   &    \checkmark    &   \checkmark     \\ \midrule
SwissProt\_Function    & $(x_{instruction},x_{multimodal\_protein})\rightarrow y_{text}$  & 411K & What is the function of this protein?                                 &    \checkmark    &   \checkmark     \\  \midrule
SwissProt\_Description & $(x_{instruction},x_{text})\rightarrow y_{prot}$  & 394K & Generate a protein matching the following description.               &    \checkmark    &    \checkmark    \\ \midrule
\multicolumn{6}{l}{\textit{\textbf{Molecule – Protein Alignment}}}                                                                                                                               \\ \midrule
BindingDB              & $(x_{instruction},x_{multimodal\_prot})\rightarrow y_{mol}$  & 335K & Generate a drug molecule binding to the target protein.             &  \checkmark      &    \checkmark    \\ \midrule
Rhea                   & $(x_{instruction},x_{multimodal\_mol})\rightarrow y_{prot}$  & 190K & Generate an enzyme that can catalyze for the given substrate.        &    \checkmark    & \checkmark  \\ \bottomrule    
\end{tabular}

\label{tab:instruction-tuning-dataset}
\end{sidewaystable}

\clearpage
\setcounter{table}{0}
\setcounter{figure}{0}




\renewcommand{\tablename}{Supplementary Table}
\renewcommand{\thetable}{\arabic{table}}
\renewcommand{\figurename}{Supplementary Figure}
\renewcommand{\thefigure}{\arabic{figure}}


\begin{flushleft}
    {\Huge \textbf{Supplementary Information for \\  Advancing Biomolecular
Understanding and Design Following Human
Instructions}}
\end{flushleft}
\vspace{1em}

\setcounter{tocdepth}{1}  
\tableofcontents
\clearpage

\section{Implementation of \modelname{}}
\subsection{Details of Multimodal Encoders}
{The molecule 2D-graph encoder, $f_m^{2D}$, is a pre-trained 5-layer GIN~\cite{DBLP:conf/iclr/XuHLJ19, DBLP:conf/iclr/HuLGZLPL20} with an output dimension of $d_m^{2D}=300$. The aggregation function of GIN can be formulated as $\mathbf{h}_v^k=\mathrm{MLP}^k((1+\epsilon^k)\mathbf{h}_v^{k-1}+\sum_{u\in\mathcal{N}_{(v)}}\mathbf{h}_u^{k-1})$, where $\mathbf{h}_v^k$ is the embedding of node $v$ at level $k$, and $\mathcal{N}_{(v)}$ represents the neighbors of node $v$.  The molecule 3D-structure encoder, $f_m^{3D}$, is Geoformer~\cite{DBLP:conf/nips/WangLWS0L23}. It is a novel geometric Transformer that integrates Interatomic Positional Encoding (IPE) to effectively model molecular structures, with an output dimension of $d_m^{3D}=512$. For protein encoding, we use pre-trained ESM2-35M~\cite{lin2022language} and SaProt-35M~\cite{su2024saprot} as the protein 1D-sequence encoder $f_p^{1D}$ and 3D-structure encoder $f_p^{3D}$, respectively, with an output dimension of 480 for both.}
\subsection{Collection of Protein Motifs}
We download data from the SwissProt database~\cite{boeckmann2003swiss} and collect all subsequences annotated as motifs. After filtering, we retain only those subsequences that appear at least twice, resulting in a total of 4712 subsequences. These subsequences are regarded as protein motifs. For ease of processing, the presence of a motif in a given protein amino acid sequence is determined by checking whether the sequence contains the corresponding subsequence.

\subsection{Training Details}
During continual pretraining, we set batch size to 32 and fix sequence length to 512 tokens. To ensure balanced training on molecule, protein, and natural language data, the sampling ratio for these three types of data is fixed at 1:1:1. In instruction-tuning, we use a batch size of 24, with the maximum sequence length set to 512. By setting hyperparameters, we control the sampling ratio of different types of instruction data. The sampling ratios for the datasets used in stage-1 and stage-2 of instruction-tuning are detailed in Supplementary Table~\ref{tab:sup:sampling-ratio}.

{The downstream task \textit{``designing protein and molecule simultaneously from descriptions''} is structured as follows: $(x_{instruction}, x_{text}) \to (y_{prot}, y_{mol})$, where $x_{instruction}$ is "\textit{Please generate the corresponding protein sequence based on the following description of the protein and generate a molecule that can interact with this protein}". $x_{text}$ is the natural language description of the protein, and $y_{prot}$, $y_{mol}$ are generated protein sequence and molecule, respectively. During training, we perform LoRA~\cite{DBLP:conf/iclr/HuSWALWWC22} fine-tuning (with hyperparameters set to $r_{lora}=32, \alpha_{lora}=32$) on \modelname{}. LoRA fine-tuning is highly efficient, updating only about 0.5\% of the model parameters.}

\subsection{Inference Settings}
For inference on downstream tasks, we load the model using the bfloat16 data format. The specific inference hyperparameters for each task are detailed in Supplementary Table~\ref{tab:sup:hyperparam}. {In the experiments of protein understanding and design following human instructions, the experiments are repeated with three different random seeds.}

\section{Baselines} In molecule captioning and description-based molecule generation tasks, the selected baselines include MolT5~\cite{DBLP:conf/emnlp/EdwardsLRHCJ22}, BioT5~\cite{DBLP:conf/emnlp/PeiZZWGWXY23}, BioT5+~\cite{DBLP:conf/acl/PeiWGLFZ00024}, MolReGPT~\cite{molregpt}, InstructMol~\cite{DBLP:journals/corr/abs-2311-16208} and ChemDFM~\cite{DBLP:journals/corr/abs-2401-14818}. {InstructMol is incapable of molecule generation.}

For protein understanding and design, target-protein-based drug design, and substrate-based enzyme design tasks, baselines are categorized into two groups. The first group comprises pre-trained models, including Mol-Instructions~\cite{DBLP:conf/iclr/FangL0LH0FC24}, InstructProtein~\cite{DBLP:conf/acl/WangZDQZLC24}, ProtT3~\cite{DBLP:conf/acl/LiuZ0ZWKC24}, BioMedGPT~\cite{DBLP:journals/corr/abs-2308-09442}, ProteinDT~\cite{DBLP:journals/corr/abs-2302-04611}, {Pocket2Mol~\cite{DBLP:conf/icml/PengLGXPM22}, AR-SBDD~\cite{DBLP:conf/nips/LuoGMP21}, TargetDiff~\cite{DBLP:conf/iclr/GuanQPS0M23}}, DrugGPT~\cite{li2023druggpt} and RFdiffusionAA~\cite{krishna2024generalized}. These models are all downloaded from their official repositories and evaluated on the test set. {In protein understanding and design task, BioMedGPT and ProtT3 cannot generate protein sequences, while ProteinDT cannot answer protein property-related questions using natural language.}
{We also train a variant of ProteinDT that incorporates motif information (ProteinDT+motif) on description-based protein design task. In this model, we add motif embeddings to the protein encoder during the first stage of training, which aligns text and protein pairs. Specifically, the protein representation is generated as: $p_p \circ [f_p(x_p)\oplus \mathbf{x}_{motif}] $, where $p_p$ is a projector, $f_p$ is the protein encoder, $x_p$ is the protein sequence, and $\mathbf{x}_{motif}$ is the motif embedding. Subsequently, we continue training the protein facilitator and protein decoder following the original procedure.} The second group consists of baseline models constructed by us using the general-purpose Large Language Model GPT-3.5. It includes three variants: zero-shot, 5-shot random, and 5-shot similarity. In the zero-shot setting, we directly pose task-specific questions to the GPT-3.5 model. In the 5-shot random setting, five examples from the training set are randomly selected as in-context demonstrations for each test entry. In the 5-shot similarity setting, the in-context learning paradigm is also adopted, but the demonstrations are required to be the five most similar examples from the training set relative to the query. The method for computing similarity depends on the data type of the input query: when query is in natural language, TF-IDF with cosine similarity is used as the text similarity measure; for protein sequence queries, MMseq2~\cite{steinegger2017mmseqs2} is employed to calculate protein similarity; and for molecule queries, molecular fingerprint similarity~\cite{rogers2010extended,bajusz2015tanimoto} is used. The specific input format for the employed GPT baseline is detailed in Supplementary Table~\ref{tab:sup:gpt}.

{For designing protein and molecule simultaneously from descriptions, baselines consist of three groups. (1) {The first group is GPT-3.5}, which includes two variants: GPT-3.5 (5-shot, random) and GPT-3.5 (5-shot, similarity). These models utilize an in-context learning paradigm. The former randomly selects 5 examples from the training set as prompts, while the latter selects the 5 most similar examples to the given query. (2) {The second group consists of fine-tuned Large Language Models}. This includes Llama~\cite{DBLP:journals/corr/abs-2307-09288}
, Mol-Instructions-molecule, Mol-Instructions-protein, BioT5, and BioT5+. Mol-Instructions-molecule and Mol-Instructions-protein are Llama-based models trained with molecule-text and protein-text instructions from the Mol-Instructions dataset~\cite{DBLP:conf/iclr/FangL0LH0FC24}, respectively. All models in this group are fine-tuned on the training set. For Llama, Mol-Instructions-molecule, and Mol-Instructions-protein, we use LoRA fine-tuning, which aligns with the approach used in \modelname{}. For BioT5 and BioT5+, we fine-tune all parameters. The number of fine-tuning steps is kept consistent across all models. (3) {The third group is a sequential ensemble approach}: GPT-3.5 (5-shot, similarity) + DrugGPT. This method combines the most competitive techniques for generating proteins from natural language (GPT-3.5 (5-shot, similarity)) and generating molecules from proteins (DrugGPT). First, GPT-3.5 (5-shot, similarity) is used to generate proteins from text, followed by the use of DrugGPT to generate molecules based on the proteins.}

\section{Datasets}


\subsection{Examples of Datasets}
We provide the example entries of continual pretraining dataset in Supplementary Table~\ref{tab:sup:pretrain-example}, and the example entries of instruction-tuning dataset in Supplementary Table~\ref{tab:sup:mt-align-example}, Supplementary Table~\ref{tab:sup:pt-align-example} and Supplementary Table~\ref{tab:sup:mp-align-example}.

\subsection{Split of Datasets}
To evaluate the model's performance, we divide the dataset used for the stage-2 instruction tuning into training and test sets. For the dataset aligning molecules with natural language, we adopt the data split from ref.~\citenum{DBLP:conf/emnlp/EdwardsLRHCJ22}. For the dataset aligning proteins with natural language, we use the training data defined in ref.~\citenum{DBLP:journals/corr/abs-2308-09442} and randomly select 3000 samples from the test set as our evaluation data. Statistics of the above two datasets are in Supplementary Table~\ref{tab:sup:dataset}. For the dataset aligning molecules with proteins, we account for the specific nature of the tasks. In the task of generating drug-like molecules for proteins, a single protein typically corresponds to multiple molecules. Conversely, in the task of generating enzymes for substrate molecules, a single substrate often corresponds to multiple proteins.  Accordingly, we split the dataset as follows: for the former task, we select 100 target proteins and their corresponding molecules as the test set. For the latter task, we select 100 target substrates and their corresponding proteins as the test set. 
The specific sizes of each dataset split are detailed in Supplementary Table~\ref{tab:sup:dataset-mp}.
{For designing protein and molecule simultaneously from descriptions task, we curate triplets from the BindingDB and SwissProt\_Description datasets, each consisting of a protein description, its corresponding protein amino acid sequence, and a binding molecule, aligned by the UniprotID. To ensure robust evaluation, we implement data filtering during the training-test split to avoid data leakage between the training and test sets. Our final training set contains 334,216 samples, with 100 protein descriptions in the test set.}

\section{{Additional Experiments}}

\subsection{{Molecule Editing}}
{we performed LoRA fine-tuning (with hyperparameters $r_{lora}=32, \alpha_{lora}=32$) on the pretrained \modelname{} model after the instruction-tuning stage. Specifically, we followed the experimental setup of ChatDrug~\cite{DBLP:conf/iclr/LiuWYW0GX24} and conducted experiments on both single-target molecule editing tasks (Task IDs 101-108) and multi-target molecule editing tasks (Task IDs 201-206).
To construct the training dataset for these tasks, we randomly selected 1,000 molecules from the retrieval database provided by ChatDrug as the original molecules. ChatDrug was then used to generate pairs of original and successfully edited molecules for each task. We applied LoRA to perform multi-task fine-tuning on \modelname{}, covering 8 single-target editing tasks and 6 multi-target editing tasks. 
The task format is $(x_{instruction}, x_{multimodal\_mol})\to y_{mol}$, with the instructions for different tasks provided in Supplementary Table~\ref{fig:response:moledit-instruct}.}

{It is worth noting that LoRA fine-tuning is a highly efficient method, as the model only needs to optimize approximately 0.5\% of the total parameters. This allows the training to be completed in just six hours on NVIDIA 3090 GPU (24GB memory). After fine-tuning, we obtained a model capable of handling various molecular editing tasks. The model was then evaluated on the test set provided by ChatDrug, using five random seeds for repeated experiments. The hit ratio (the proportion of successful edits) was used as the evaluation metric.}

{In Supplementary Tables~\ref{tab:response:moledit-single} and \ref{tab:response:moledit-multi}, we report the experimental results for single-target and multi-target editing tasks, respectively. Each editing task includes two different threshold settings: a loose threshold and a strict threshold, resulting in a total of 28 tasks. The average performance of the model across all tasks is also provided. It is evident that \modelname{} demonstrates the best overall performance. In both single-target and multi-target editing tasks, \modelname{} achieves the highest average performance and outperforms the other models in more than half (15/28) of the tasks. More importantly, we also observe that \modelname{} shows particularly significant performance improvement in the more challenging multi-target molecule editing tasks, with a 17.8\% increase. This result underscores \modelname{}'s ability not only to improve editing accuracy but also to enhance efficiency when handling complex molecule editing tasks, further showcasing its strengths in the molecular editing domain. We have also updated the code repository with the training and inference code for LoRA fine-tuning on molecular editing tasks, along with the associated data and model weights.}

{Through this molecule editing experiment, we demonstrate that \modelname{} can efficiently adapt to various downstream tasks via LoRA fine-tuning. This flexibility arises from \modelname{} being a unified natural language-molecule-protein model, trained on large-scale and diverse datasets, enabling excellent any-to-any alignment. As a result, \modelname{} can tackle a wide range of complex tasks in the biomolecular domain, providing a strong foundation for further research and applications. This highlights \modelname{}'s tremendous potential as a foundational model for biomolecular tasks.}

\subsection{{Designing Molecules based on Pairs of Target Proteins}}

{Generalizing to multiple molecule-protein alignments is indeed an exciting area of exploration. As mentioned in the Discussion Section, due to resource constraints, we did not include all potential biomolecular tasks in the current study. However, the model architecture and training framework we propose are designed to be highly generalizable and scalable, allowing for easy adaptation to a variety of tasks.}

{To explore how the model could generalize to multiple molecule-protein alignment tasks—such as generating molecules that bind or do not bind to several proteins—we found that LoRA fine-tuning offers a simple yet highly effective solution. LoRA only requires optimizing 0.5\% of the model parameters, and it can be completed with high computational efficiency on 24GB NVIDIA 3090 GPU.
Specifically, we applied LoRA fine-tuning to conduct experiments on generating molecules that bind to a pair of proteins. First, we collected data on protein pairs and molecules that bind to both of them simultaneously, and divided the data into training and test sets. The training set contains 102,075 protein pairs-molecule binding triplets, and the test set includes 30 pairs of protein targets. We adopted the same training paradigm as in the instruction-tuning phase: $(x_{instruction}, x_{multimodal\_prot1}, x_{multimodal\_prot2}) \to y_{mol}$, where $x_{instruction}$ is \textit{``Generate a drug molecule that binds to these target proteins"}, $x_{multimodal\_prot1}$ and $x_{multimodal\_prot2}$ are the two target proteins, and $y_{mol}$ is the generated molecule. The hyperparameters for LoRA were set to $r=32$ and $\alpha =32$, and the fine-tuning process required only about half an hour.}

{In the experiment, we used Vina Score, High Affinity, and Success Rate as evaluation metrics. For calculating High Affinity and Success Rate, we required that the generated molecules satisfy the conditions on both target proteins. Since this is a very novel task and no specialized models have been designed for it, we compared \modelname{} with the general model GPT-3.5 and its in-context learning variant. The experimental results in Supplementary Figure~\ref{fig:response:multip2m} show that \modelname{} significantly outperforms the baseline models. This highlights the model’s ability to quickly adapt to new tasks through LoRA-based instruction-tuning, showcasing its excellent scalability and generalization. Our findings suggest that \modelname{}’s any-to-any alignment allows it to handle complex tasks, further underscoring its potential for broad applications in biomolecular research.}

\subsection{{Ablation Analysis}}

\paragraph{{Analysis of Training Strategy.}}
{
We analyze the impact of different training strategies. Supplementary Figure~\ref{fig:ablation}a compares \modelname{} with two variants: one that retains continual pretraining and stage-2 instruction-tuning but removes stage-1 instruction-tuning ("w/ continual-pretraining, w/o stage-1 instruction-tuning"), and another that only retains stage-2 instruction-tuning, removing both continual pretraining and stage-1 instruction-tuning ("w/o continual-pretraining, w/o stage-1 instruction-tuning"). Our findings are as follows:}

{For tasks related to molecule understanding and design (e.g., molecule captioning and description-based generation), continual pretraining has a minor impact, while stage-1 instruction-tuning is more crucial. In molecule captioning, removing stage-1 instruction-tuning results in a 7.3\% performance drop, with an additional 1.2\% decrease after removing continual pretraining. In molecule generation, removing stage-1 instruction-tuning leads to a 12.4\% drop in accuracy, with only a 1.1\% further decrease after removing continual pretraining. We attribute this to the role of molecular IUPAC name data in stage-1 instruction-tuning, which aligns molecular structure with natural language. For description-based protein generation, continual pretraining plays a more significant role, while stage-1 instruction-tuning has a smaller impact. Removing stage-1 instruction-tuning decreases the Identity by 3.6\%, while further removal of continual pretraining results in an 8.9\% drop. This suggests that the evolutionary information captured during continual pretraining, particularly from broad protein sequence datasets, enhances the quality of generated proteins.}

\paragraph{{Analysis of Multimodal Data Integration.}}
{Next, we explore the impact of incorporating multimodal data on model performance. As shown in Supplementary Figure~\ref{fig:ablation}b, we compare \modelname{} and its variants on molecule captioning and protein function answering tasks. Variants include the removal of the 2D encoder (w/o 2D encoder) and 3D encoder (w/o 3D encoder) for molecular data, as well as the removal of the 1D encoder (w/o 1D encoder) and 3D encoder (w/o 3D encoder) for protein data. Additionally, we consider the removal of motif prompts (w/o motif) and the entire Motif-Guided Multimodal Feature Extraction Module (w/o multimodal). \modelname{} and its variants here exclude both continual pretraining and stage-1 instruction-tuning.}

{The results reveal that removing the Motif-Guided Multimodal Feature Extraction Module significantly decreases performance, with average drops of 4.7\% and 7.1\% on molecular and protein tasks, respectively. This highlights the importance of multimodal feature extraction, which integrates information from different modalities to provide more comprehensive and accurate representations. Further analysis shows that 3D data has minimal impact on molecular tasks, but is more crucial for protein tasks, where the removal of the 3D encoder leads to a 4.2\% decline. This suggests that 3D structures are essential for identifying protein functions. Overall, multimodal data enhances performance in biomolecular tasks, emphasizing its importance in these domains.}

\section{Additional Analysis}

\subsection{{Analysis of Motifs in Biomolecules}}
{In Supplementary Figures~\ref{fig:response:motif-mol} and \ref{fig:response:motif-prot}, we provide examples of shared motifs in both molecules and proteins, along with their corresponding textual descriptions. By examining these examples, we observe a clear pattern: when molecules or proteins share common motifs, their associated natural language descriptions also show notable similarities. This observation suggests a strong correlation between motif information and natural language descriptions, highlighting the critical role motifs play in capturing the functional properties of biomolecules. As such, the incorporation of motif-based features into the model improves its ability to capture these similarities and better understand molecular functions, ultimately enhancing model performance.}

\subsection{{Novelty of the Generated Biomolecules}}
{We compared two aspects of novelty~\cite{du2024machine}: (1) the proportion of generated drugs or enzymes that do not appear in the reference set, and (2) the proportion that does not appear in the training set.
The results are shown in Supplementary Table~\ref{tab:response:novel-drug} and Supplementary Table~\ref{tab:response:novel-enzyme}.
In the drug generation task, our model achieved 100\% novelty with respect to the reference set, meaning that all the generated molecules were novel and did not appear in the reference data. This highlights the ability of our approach to expand the known drug space for the given targets. Additionally, when compared to the training set, our method reached a 92.9\% novelty rate, representing a 1.5\% improvement over the baseline method, DrugGPT.
In the enzyme generation task, our model achieved novelty scores of 99.8\% compared to the reference set, and 98.0\% compared to the training set. While these novelty scores are slightly lower than some baseline methods, they still surpass GPT-3.5 (5-shot, similarity). More importantly, the enzymes generated by our model show substantial improvements in substrate-enzyme interaction, as illustrated in Figure 3d of the Manuscript.
These results underscore the potential of our approach to discover novel drugs and enzymes that have not been previously reported or recorded, emphasizing its capability to contribute to the exploration of new biomolecular spaces.}

\subsection{Case Analysis for Experimental Results of Understanding and Designing Molecules}
For the task of description-based molecule generation, several case examples of the ground truth and generation are presented in Supplementary Figure~\ref{fig:sup-mol2text-demo}. Overall, \modelname{} demonstrates a relatively accurate analysis of molecular structure, function, origin, etc.
In the task of description-based molecule generation, \modelname{} is capable of generating molecules that are completely consistent with the ground truth in certain cases, such as molecules PubChem-CID-5281294 and PubChem-CID-31284 shown in Supplementary Figure~\ref{fig:sup-text2mol-demo}. This demonstrates the strong molecular design capability of \modelname{}. However, in some other cases, the molecules generated by \modelname{} show some discrepancies with the ground truth. Our analysis suggests that this may be due to inadequate handling of certain functional groups. For example, for PubChem-CID-118429016 and PubChem-CID-123953, the model omits certain functional groups (a hydroxyl group and a phosphate group, respectively). For PubChem-CID-179394, the model generates a chemically atypical P(O)(O)(O)O group. Overall, \modelname{} exhibits a high level of accuracy in molecule generation tasks and shows potential for applications in fields such as drug discovery, and further optimization may be required in practical applications.

\subsection{Case Analysis for Experimental Results of Understanding and Designing Proteins}
In the protein function answering task, \modelname{} generates results that closely resemble the ground truth for certain cases, such as Q9NRY2 and P73070 in Supplementary Table~\ref{tab:sup:protein2text-demo}. Furthermore, we observe that in some cases, the generated descriptions tend to be more detailed.
For example, in the case of Q9Y2G3, the generated description includes detailed information on vesicle formation, lipid signal molecule uptake, and the establishment of the thrombopoietin gradient in platelets.
Similarly, for Q9FY89, the generated description provides a more comprehensive explanation of the formation of multivesicular bodies (MVBs), specifically describing the formation mechanism of intraluminal vesicles (ILVs) within MVBs, including the invagination and scission of the endosomal membrane. It further elaborates on the function of MVBs, such as transporting their contents to lysosomes for the degradation of membrane proteins, receptors, lysosomal enzymes, and lipids.
For P0CP67, the generated description offers more detailed and in-depth functional information, identifying the specific targets of the protein's action (e.g., components of AP-1, c-Jun, and ATF2) and potential biological processes involved (e.g., regulation of circadian clock). 
Although \modelname{} may provide researchers with deeper insights, further experimental validation is necessary to confirm these findings.

For the task of text-based protein generation, we present two examples in Supplementary Figure~\ref{fig:sup-text2protein-demo}. For the ground truth proteins, we utilize the protein structures predicted by AlphaFold Protein Structure Database~\cite{varadi2022alphafold}. For the generated protein sequences, we predict their structures using ColabFold~\cite{mirdita2022colabfold}. Besides sequence similarity metrics Identity, Alignment, and BLOSUM Substitution, we also compare structural similarity metrics: TM-Score~\cite{zhang2004scoring} and LDDT~\cite{mariani2013lddt}. The results demonstrate that \modelname{} is capable of de-novo design of proteins, with the designed proteins exhibiting a high degree of structural similarity to ground truth. This suggests that \modelname{} holds significant potential in designing proteins tailored to specific functional descriptions, acting as an effective copilot to assist researchers in protein design.

\clearpage

\begin{table}[t]
\centering
\caption{Sampling ratios of different types of instruction data during instruction-tuning. Note that in practice, the sampling ratios are scaled proportionally to ensure that the sum of the ratios for all data equals 1.}
\begin{tabular}{llcc}
\toprule
\multirow{2}{*}{Sub-dataset} & \multirow{2}{*}{Task type} & \multicolumn{2}{c}{Sampling ratio} \\ \cmidrule{3-4} 
 &  & stage-1 & stage-2 \\ \hline
\multicolumn{4}{l}{\textit{\textbf{Molecule - Natural Language Alignment}}}    \\ \midrule
PubChem                & $(x_{instruction},x_{multimodal\_mol})\rightarrow y_{text}$    &  0.1    &   -     \\ \midrule
PubChem                & $(x_{instruction},x_{text})\rightarrow y_{mol}$ &   0.1    &   -     \\ \midrule
ChEBI                  & $(x_{instruction},x_{multimodal\_mol})\rightarrow y_{text}$  &    0.001    &    0.1    \\ \midrule
ChEBI                  & $(x_{instruction},x_{text})\rightarrow y_{mol}$  &    0.001    &     0.1   \\ \midrule
\multicolumn{4}{l}{\textit{\textbf{Protein - Natural Language Alignment}}}                                                                                                                       \\ \midrule
TrEMBL\_Name           & $(x_{instruction},x_{multimodal\_protein})\rightarrow y_{text}$                      &    0.05    &     -   \\ \midrule
TrEMBL\_Family         & $(x_{instruction},x_{multimodal\_protein})\rightarrow y_{text}$            &   0.05     &    -    \\ \midrule
TrEMBL\_Locaction      & $(x_{instruction},x_{multimodal\_protein})\rightarrow y_{text}$         &     0.05   &    -    \\ \midrule
TrEMBL\_Function       & $(x_{instruction},x_{multimodal\_protein})\rightarrow y_{text}$           &    0.05   &   -     \\ \midrule
TrEMBL\_Description    & $(x_{instruction},x_{text})\rightarrow y_{prot}$                &     0.1   &   -     \\ \midrule
SwissProt\_Name        & $(x_{instruction},x_{multimodal\_protein})\rightarrow y_{text}$           &   0.05     &     0.1   \\ \midrule
SwissProt\_Family      & $(x_{instruction},x_{multimodal\_protein})\rightarrow y_{text}$      &    0.05    &   0.1 \\ \midrule
SwissProt\_Location    & $(x_{instruction},x_{multimodal\_protein})\rightarrow y_{text}$  &    0.05    &  0.1   \\ \midrule
SwissProt\_Function    & $(x_{instruction},x_{multimodal\_protein})\rightarrow y_{text}$    &    0.05   &   0.1 \\  \midrule
SwissProt\_Description & $(x_{instruction},x_{text})\rightarrow y_{prot}$     &    0.1   &   0.2    \\ \midrule
\multicolumn{4}{l}{\textit{\textbf{Molecule – Protein Alignment}}}   \\ \midrule
BindingDB              & $(x_{instruction},x_{multimodal\_prot})\rightarrow y_{mol}$        &  0.05      &    0.1    \\ \midrule
Rhea                   & $(x_{instruction},x_{multimodal\_mol})\rightarrow y_{prot}$      &    0.05    & 0.1  \\ \bottomrule  
\end{tabular}
\label{tab:sup:sampling-ratio}
\end{table}

\begin{table}[t]
\centering
\caption{Inference hyperparameters for downstream tasks.}
\begin{tabular}{ll}
\toprule
Downstream Task & Inference Hyperparameter \\ \midrule
Molecule captioning & $num\_beams=5$ \\
Description-based molecule generation & $num\_beams=5$ \\ \midrule
Protein property answering & $top\_p=0.1, temperature=1$ \\
Description-based protein generation & $top\_p=0.9, temperature=0.8$ \\ \midrule
Target protein-based drug discovery & $top\_p=1, temperature=1$ \\
Target substrate-based enzyme design & $top\_p=0.9, temperature=0.8$ \\ \midrule
Designing protein and molecule from descriptions & $top\_p=0.7, temperature=1$ \\ \bottomrule
\end{tabular}
\label{tab:sup:hyperparam}
\end{table}

\begin{table}[]
\centering
\caption{In-context learning examples of the GPT baseline across different tasks, Where \textit{XX} represents data-specific natural language descriptions, protein sequences, or molecular sequences. In the few-shot setting, the input to the GPT model consists of a template, in-context demonstrations, and a question; in the zero-shot setting, the input consists of a template and a question.}
\scalebox{0.8}{{
\begin{tabular}{p{9cm}p{7cm}p{3cm}}
\toprule
Template & In-context Demonstration & Question \\ \midrule
\multicolumn{3}{l}{\textbf{\textit{Protein property answering}}}  \\ \midrule
You are a biologist. Given the protein sequence, your task is to generate a family of this protein using your experienced knowledge. & Please strictly follow the format, no other information can be provided. Protein sequence: \textit{XX}; Protein family: \textit{XX}. ... Protein sequence: \textit{XX}; Protein family: \textit{XX} & Protein sequence: \textit{XX}; Protein family: \\ \midrule
You are a biologist. Given the protein sequence, your task is to generate a subcellular localization of this protein using your experienced knowledge. & Please strictly follow the format, no other information can be provided. Protein sequence: \textit{XX}; Protein subcellular localization: \textit{XX}. ... Protein sequence: \textit{XX}; Protein subcellular localization: \textit{XX} & Protein sequence: \textit{XX}; Protein subcellular localization: \\ \midrule
You are a biologist. Given the protein sequence, your task is to generate a name of this protein using your experienced knowledge. & Please strictly follow the format, no other information can be provided. Protein sequence: \textit{XX}; Protein name: \textit{XX}. ... Protein sequence: \textit{XX}; Protein name: \textit{XX} & Protein sequence: \textit{XX}; Protein name: \\ \midrule
You are a biologist. Given the protein sequence, your task is to generate a function of this protein using your experienced knowledge. & Please strictly follow the format, no other information can be provided. Protein sequence: \textit{XX}; Protein function: \textit{XX}. ... Protein sequence: \textit{XX}; Protein function: \textit{XX} & Protein sequence: \textit{XX}; Protein function: \\ \midrule
\multicolumn{3}{l}{\textbf{\textit{Description-based protein generation}}}   \\ \midrule
You are a biologist. Given the protein description, your task is to design a new protein matching the description using your experienced knowledge. You MUST reply using a sequence of the capitalized initial letters of 20 amino acids and DO NOT reply with others. & Please strictly follow the format, no other information can be provided. Protein description: \textit{XX}, Protein sequence: \textit{XX}. ... Protein description: \textit{XX}, Protein sequence: \textit{XX}. & Protein description: \textit{XX}, Protein sequence: \\ \midrule
\multicolumn{3}{l}{\textbf{\textit{Target protein-based drug discovery}}} \\ \midrule
You are a biologist. Given the protein, your task is to design a drug molecule binding to this protein using your experienced knowledge. You should only reply with SMILES and DO NOT reply with others. & Please strictly follow the format, no other information can be provided. Protein sequence: \textit{XX}; Molecule: \textit{XX}. ... Protein sequence: \textit{XX}; Molecule: \textit{XX}. & Protein sequence: \textit{XX}; Molecule: \\ \midrule
\multicolumn{3}{l}{\textbf{\textit{Target substrate-based enzyme design}}}  \\ \midrule
You are a biologist. Given the molecule, your task is to design an enzyme protein that can catalyze for this substrate using your experienced knowledge. You MUST reply using a sequence of the capitalized initial letters of 20 amino acids and DO NOT reply with others. & Please strictly follow the format, no other information can be provided. Molecule: \textit{XX}; Protein sequence: \textit{XX}. ... Molecule: \textit{XX}; Protein sequence: \textit{XX}. & Molecule: \textit{XX}; Protein sequence: \\ \midrule
\multicolumn{3}{l}{\textbf{\textit{Designing protein and molecule simultaneously from descriptions}}}  \\ \midrule
You are a biologist. Given the protein description, your task is to design a new protein and a molecule that can bind to this protein using your experienced knowledge. Please respond only with the amino acid sequence using a sequence of the capitalized initial letters of 20 amino acids and SMILES representation of molecule, separated by `,', and DO NOT reply with others. & Please strictly follow the format, no other information can be provided. Protein Description: \textit{XX}; Protein and Molecule: \textit{XX}, \textit{XX} . ... Protein Description: \textit{XX}; Protein and Molecule: \textit{XX}, \textit{XX} . & Protein Description: \textit{XX}; Protein and Molecule:  \\ \bottomrule
\end{tabular}
}}
\label{tab:sup:gpt}
\end{table}

\begin{table}[t]
\centering
\small
\caption{Examples of the continual pertaining data.}
\scalebox{0.9}{{
\begin{tabular}{lp{14cm}}
\toprule
Data type        & Data               \\ \midrule
Molecule         & \texttt{[C][C][C][C][S][P][=Branch1][C][=O][Branch1][=Branch1][S][C][C] [C][C][S][C][C][C][C]}                                                    \\ \midrule
Protein          & \texttt{MLSKNNNRELKRKMEEKQDRFTIKKLSVGVASVLLGSFIMGTQAVQTAHASDDNTEDATVNSAQ NTTMEQVVPLTASTS}                                                                          \\ \midrule
Natural language & Antiphospholipid syndrome (APS) and heparin-induced thrombocytopenia (HIT) are thrombotic disorders due to specific autoimmune-mediated antibodies. Catastrophic APS (CAPS), also known as Asherman's syndrome, is a life-threatening severe form of APS. Diagnostic criteria for CAPS include the development of a thrombotic event of three or more organs in less than a week with the presence of antiphospholipid antibodies and ... \\ \bottomrule
\end{tabular}
}}
\label{tab:sup:pretrain-example}
\end{table}

\begin{table}[t]
\centering
\caption{Examples of the instruction-tuning data on molecule-natural language alignment.}
\scalebox{0.9}{{
\begin{tabular}{lp{3cm}p{6cm}p{6cm}}
\toprule
Dataset                  & Instruction        & Input            & Output            \\ \midrule
\multirow{2}{*}[-3em]{PubChem} & Give the IUPAC name of the following molecule.                     &  \texttt{[C][=C][Branch1][C][F][C] [Branch1][C][O][C][C][C]}                                            & 2-fluorohex-1-en-3-ol                               \\ \cmidrule{2-4}
                         & Generate a molecule in SELFIES that fits the provided IUPAC name.  & 2-fluorohex-1-en-3-ol                                        & \texttt{[C][=C][Branch1][C][F][C] [Branch1][C][O][C][C][C]}                                         \\ \midrule
\multirow{2}{*}[-7.5em]{ChEBI}   & Provide a caption for the molecule below.                          & \texttt{[C][C][C][C][C][O][C] [=Branch1][C][=O][C][=C][C] [=C][C][=C][Ring1][=Branch1] [C][=Branch1][C][=O][O]}         & The molecule is a phthalic acid monoester obtained by formal condensation of one of the carboxy groups of phthalic acid with the hydroxy group of pentanol. It has a role as a xenobiotic metabolite, an anti-estrogen and a rat metabolite. It derives from a pentan-1-ol. \\ \cmidrule{2-4}
                         & Generate a molecule in SELFIES that fits the provided description. & The molecule is a phthalic acid monoester obtained by formal condensation of one of the carboxy groups of phthalic acid with the hydroxy group of pentanol. It has a role as a xenobiotic metabolite, an anti-estrogen and a rat metabolite. It derives from a pentan-1-ol. & \texttt{[C][C][C][C][C][O][C] [=Branch1][C][=O][C][=C] [C][=C][C][=C][Ring1] [=Branch1][C][=Branch1][C] [=O][O]}   \\ \bottomrule
\end{tabular}
}}
\label{tab:sup:mt-align-example}
\end{table}

\begin{table}[t]
\centering
\small
\caption{Examples of the instruction-tuning data on protein-natural language alignment.}
\scalebox{0.85}{{
\begin{tabular}{lp{2.5cm}p{6.5cm}p{5cm}}
\toprule
Dataset                  & Instruction                                                        & Input                            & Output                 \\ \midrule
TrEMBL\_Name           & What is the official name of this protein?               & \texttt{MFRRGYAKYCFDNGISIYDISLSMGHSN INTTVSYINKNSDDISIYKIFNQI}                                       & Tyr recombinase domain-containing protein                                                 \\ \midrule
TrEMBL\_Family         & What is the protein family that this protein belongs to? & \texttt{MRKLMALCALAGVVLVTGCNTMAGAGKD IEKGGEKVQGAAESVKQKM}                                                                                     & Belongs to the EcnA/EcnB lipoprotein family.                                         \\ \midrule
TrEMBL\_Locaction      & What is the subcellular localization of this protein?        & \texttt{LNMAENSCIDRCVSKYWQVTNLVGQLLG NNQPPM}           & Mitochondrion inner membrane Peripheral membrane protein Intermembrane side             \\ \midrule
TrEMBL\_Function       & What is the function of this protein?                    & \texttt{MFDQATKLHFRGARIWLAVVEDLMAKGM RHAENVRNTLNILSTCSLL}                                                                         & Hydrolyzes acetyl esters in homogalacturonan regions of pectin. In type I primary cell wall, galacturonic acid residues of pectin can be acetylated at the O-2 and O-3 positions. Decreasing the degree of acetylation of pectin gels in vitro alters their physical properties. \\ \midrule
TrEMBL\_Description    & Generate a protein matching the following description.   & The protein is Phospholipid scramblase. It belongs toPhospholipid scramblase family. FUNCTION: It may mediate accelerated ATP-independent bidirectional transbilayer migration of phospholipids upon binding calcium ions that results in a loss of phospholipid asymmetry in the plasma membrane.                                                                          & \texttt{MQEMLTDADTFSATFPLNLDVN VKAGLLAATFLIDFLYFEDE}                                                                            \\ \midrule
SwissProt\_Name        & What is the official name of this protein?               & \texttt{DCCRKPFRKHCWDCTAGTPYYGYSTRNI FGCTC}            & Mytimycin.                                                                                       \\ \midrule
SwissProt\_Family      & What is the protein family that this protein belongs to? & \texttt{TRSGGACNSHNQCCDDFCSTATSTCV}                                          & Belongs to the conotoxin O1 superfamily.                                                                               \\ \midrule
SwissProt\_Location    & What is the subcellular localization of this protein?        & \texttt{MRIAKIGVIALFLFMALGGIGGVMLAGY TFILRAG}                                                          & Cell inner membrane; Single-pass membrane protein.                                                                    \\ \midrule
SwissProt\_Function    & What is the function of this protein?                    & \texttt{GKIPIGAIKKAGKAIGKGLRAVNIASTA HDVYTFFKPKKRH}                                                                     & Has antibacterial activity against Gram-positive and Gram-negative bacteria.                                    \\ \midrule
SwissProt\_Description & Generate a protein matching the following description.   & The protein is Photosystem II reaction center protein M, PSII-M. The protein is located in the plastid, specifically on the chloroplast thylakoid membrane. It is a single-pass membrane protein. It belongs to the PsbM family. The protein is one of the components of the core complex of photosystem II (PSII). PSII is a light-driven water:plastoquinone oxidoreductase that uses light energy to abstract electrons from H(2)O, generating O(2) and a proton gradient subsequently used for ATP formation. It consists of a core antenna complex that captures photons, and an electron transfer chain that converts photonic excitation into a charge separation. This subunit is found at the monomer-monomer interface. & \texttt{MEVNILAFIATALFILVPTAFL LIIYVKTVSQNN}          \\ \bottomrule
\end{tabular}
}}
\label{tab:sup:pt-align-example}
\end{table}

\begin{table}[t]
\centering
\caption{Examples of the instruction-tuning data on molecule-protein alignment.}
\scalebox{0.85}{{
\begin{tabular}{lp{3cm}p{6cm}p{6cm}}
\toprule
Dataset                  & Instruction                                                        & Input                            & Output                 \\ \midrule
BindingDB & Generate a drug molecule binding to the target protein.       & \texttt{MLRQIIGQAKKHPSLIPLFVFIGT GATGATLYLLRLALFNPDVCWDRN NPEPWNKLGPNDQYKFYSVNVDYS KLKKERPDF} & \texttt{[C][O][C][=C][C][Branch2] [Ring1][O][C][O][C][=Branch1] [C][=O][C][=C][C][=C][O][C] [Branch1][C][C][Branch1][C][C] [C][=C][C][Ring1][Branch2][=C] [Ring1][N][=C][C][Branch1] [Ring1][O][C][=C][Ring2] [Ring1][Branch2][O][C]} \\ \midrule
Rhea      & Generate an enzyme that can catalyze for the given substrate. & \texttt{[O][=C][Branch1][C][O-1] [C][=Branch1][C][=O][C][O]}                                & \texttt{DLFHAQRGHGNLTQTLTDYMPYIGHIQI SQVPSRHEPDSDGEINYPFIFHTIAKLG YKGWVGCEYTPRGKTQLTV}              \\ \bottomrule                                                                                                                            
\end{tabular}
}}
\label{tab:sup:mp-align-example}
\end{table}

\begin{table}[t]
\centering
\caption{Statistics of datasets for molecule-natural language alignment and protein-natural language alignment.}
\begin{tabular}{lcc}
\toprule
Dataset               & Training Set & Test Set \\ \midrule
ChEBI                 & 26,407        & 3,300     \\
SwissProt\_Name        & 455,583       & 3,000     \\
SwissProt\_Family      & 370,642       & 3,000     \\
SwissProt\_Location    & 275,740       & 3,000     \\
SwissProt\_Function    & 411,064       & 3,000     \\
SwissProt\_Description & 393,818       & 3,000     \\ \bottomrule
\end{tabular}
\label{tab:sup:dataset}
\end{table}

\begin{table}[t]
\centering
\caption{Statistics of datasets for molecule-protein alignment. \# of Entries and \# of Targets denote the number of data entries and the number of targets, respectively.}
\begin{tabular}{lcccc}
\toprule
\multirow{2}{*}{Dataset} & \multicolumn{2}{c}{Training Set} & \multicolumn{2}{c}{Test Set} \\ \cmidrule{2-5}
                         & \# of Entries     & \# of Targets    & \# of Entries  & \# of Targets  \\ \midrule
BindingDB                & 335,450           & 2,033          & 1,612          & 100          \\
Rhea                     & 190,206           & 926           & 2,207          & 100          \\ \bottomrule
\end{tabular}
\label{tab:sup:dataset-mp}
\end{table}

\begin{figure*}[!t]
\centering
\includegraphics[width=\linewidth]{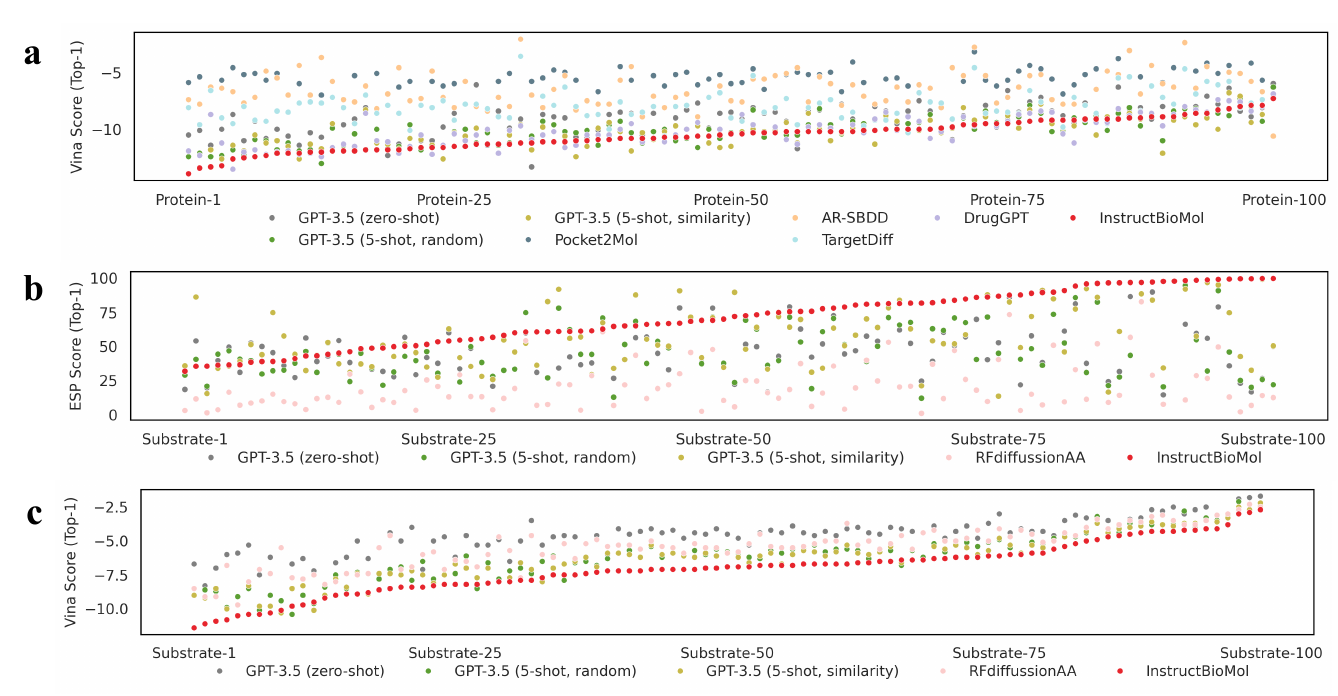}
\caption{
\textbf{a}, Performance comparison on protein-based drug discovery task. Comparing \modelname{} and baselines in terms of the Top-1 Vina Score against all proteins in the test set.
\textbf{b, c}, Performance comparison on substrate-based enzyme design task. Comparing \modelname{} and baselines in terms of the Top-1 ESP Score (\textbf{b}) and Top-1 Vina Score (\textbf{c}) against all substrates in the test set.}
\label{fig:sup-vina-esp-scatter}
\end{figure*}

\begin{figure*}[t]
\centering
\includegraphics[width=\linewidth]{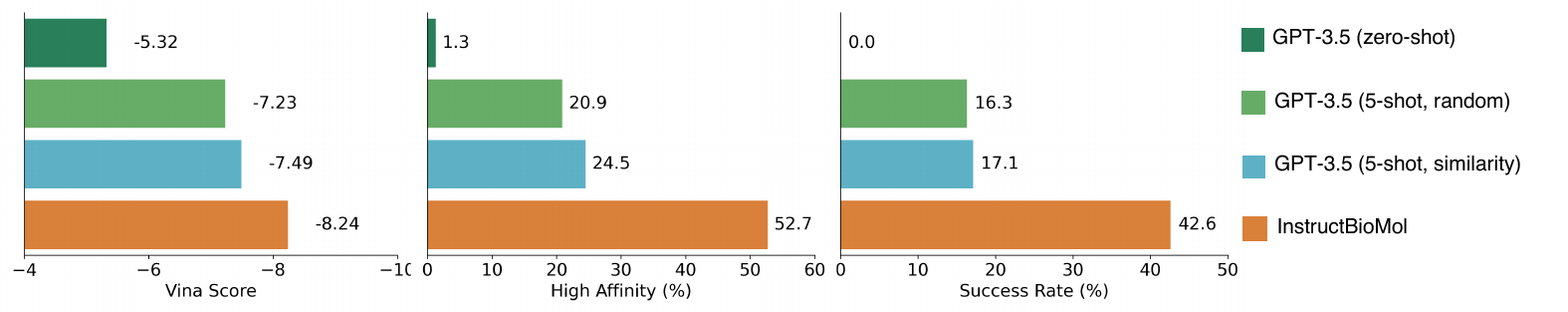}
\caption{{Results for designing molecules binding to pairs of proteins.}}
\label{fig:response:multip2m}
\end{figure*}

\begin{figure*}[!t]
\centering
\includegraphics[width=\linewidth]{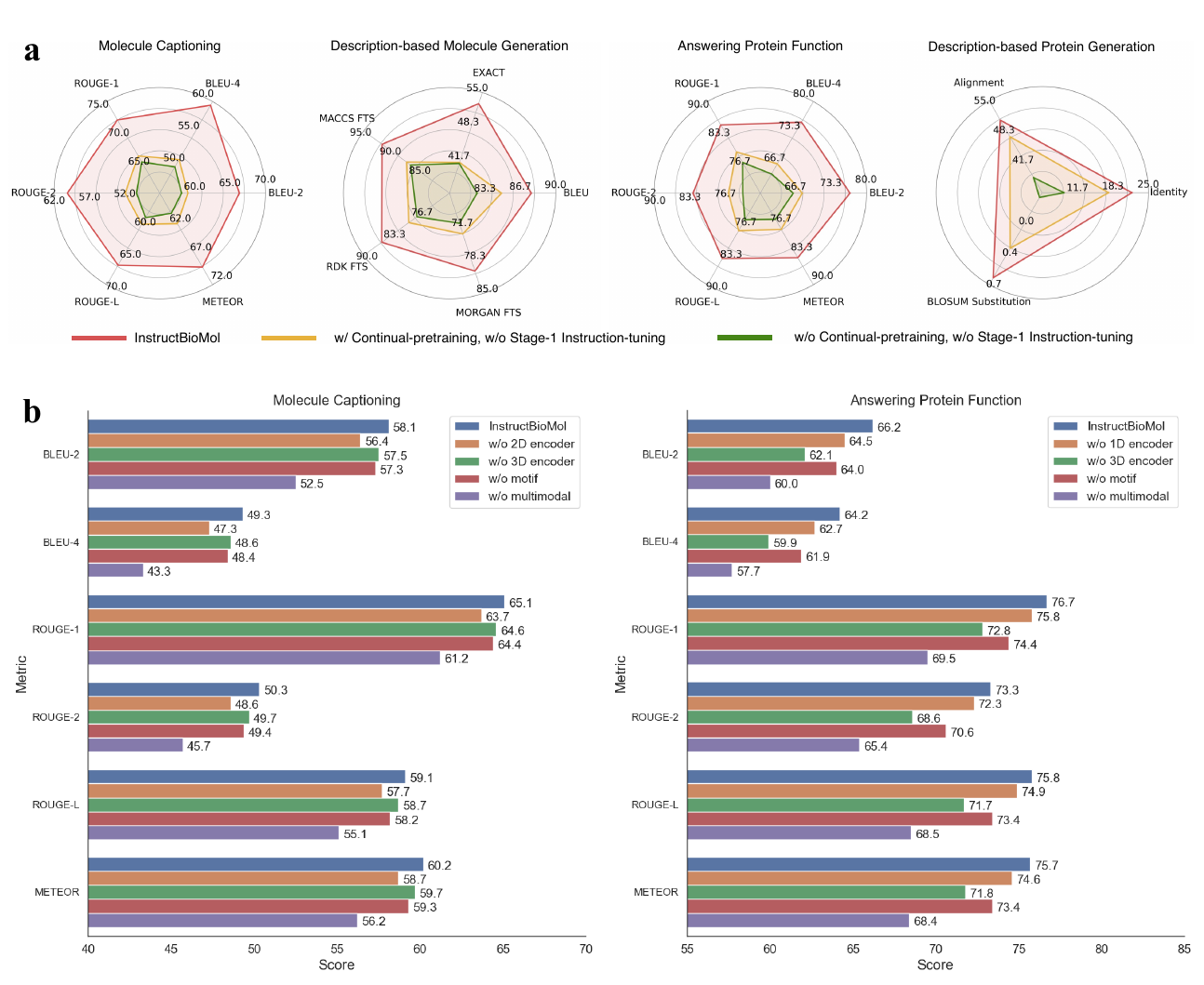}
\caption{{\textbf{Results of ablation analysis on the impact of training strategy and the multimodal data integration.} \textbf{a}, The impact of removing specific training stages is evaluated across four tasks: molecule captioning, description-based molecule generation, protein function answering, and description-based protein generation. The models compared include \modelname{} and its two variants: one variant retains the continual-pretraining stage but removes the stage-1 instruction-tuning (w/ continual-pretraining, w/o stage-1 instruction-tuning), while the other variant removes both the continual-pretraining stage and the stage-1 instruction-tuning (w/o continual-pretraining, w/o stage-1 instruction-tuning). \textbf{b}, The comparison of multimodal data integration, involving \modelname{} and its various variants: w/o 1D encoder, w/o 2D encoder, and w/o 3D encoder, which correspond to the removal of specific modality inputs. Additionally, w/o motif denotes the removal of motif prompts, and w/o multimodal represents the removal of the entire Motif-Guided Multimodal Feature Extraction Module.}}
\label{fig:ablation}
\end{figure*}

\begin{table}[t]
\centering
\caption{{Instructions for molecule editing tasks.}}
\scalebox{0.95}{{
\begin{tabular}{ll}
\toprule
Task ID & Instruction \\ \midrule
101 & Make the molecule more soluble in water. \\
102 & Make the molecule less soluble in water. \\
103 & Make the molecule more like a drug. \\
104 & Make the molecule less like a drug. \\
105 & Make the molecule higher permeability. \\
106 & Make the molecule lower permeability. \\
107 & Make the molecule with more hydrogen bond acceptors. \\
108 & Make the molecule with more hydrogen bond donors. \\ \midrule
201 & Make the molecule more soluble in water and more hydrogen bond acceptors. \\
202 & Make the molecule less soluble in water and more hydrogen bond acceptors. \\
203 & Make the molecule more soluble in water and more hydrogen bond donors. \\
204 & Make the molecule less soluble in water and more hydrogen bond donors. \\
205 & Make the molecule more soluble in water and higher permeability. \\
206 & Make the molecule more soluble in water and lower permeability. \\ \bottomrule
\end{tabular}
}}
\label{fig:response:moledit-instruct}
\end{table}

\begin{table}[t]
\centering
\caption{{The hit ratio results for 16 single-objective molecule editing tasks, along with the average performance across all tasks. For \modelname{}, we present the mean and standard deviation across five random seeds. The best-performing results are highlighted in \textbf{bold}.} The number of experimental replicates $n$=5.}  
\label{tab:response:moledit-single} 
\scalebox{0.8}{
\begin{tabular}{llcccccc}
\toprule
\multirow{2}{*}{Single Target Property} & \multirow{2}{*}{$\Delta$} & \multicolumn{2}{c}{MoleculeSTM} & \multicolumn{3}{c}{ChatDrug} & \multirow{2}{*}{\modelname{}} \\ \cmidrule(r){3-4} \cmidrule(r){5-7}
 &  & \multicolumn{1}{c}{SMILES} & \multicolumn{1}{c}{Graph} & \multicolumn{1}{c}{GALACTICA} & \multicolumn{1}{c}{Llama2} & \multicolumn{1}{c}{Turbo} & \multicolumn{1}{c}{} \\ \midrule
\multirow{2}{*}{101 more soluble in water} & 0 & 61.87$\pm$2.67 & 67.86$\pm$3.46 & 83.32$\pm$1.13 & 42.88$\pm$1.83 & \textbf{94.13$\pm$1.04} & 88.40$\pm$1.51 \\
 & 0.5 & 49.02$\pm$1.84 & 54.44$\pm$3.99 & 78.20$\pm$1.85 & 30.31$\pm$1.91 & \textbf{88.67$\pm$0.95} & 78.20$\pm$2.11 \\ \midrule
\multirow{2}{*}{102 less soluble in water} & 0 & 52.71$\pm$1.67 & 64.79$\pm$2.76 & 72.41$\pm$4.44 & 47.89$\pm$2.05 & \textbf{96.86$\pm$1.10} & 90.80$\pm$1.89 \\
 & 0.5 & 30.47$\pm$3.26 & 47.09$\pm$3.42 & 61.47$\pm$3.55 & 34.76$\pm$2.46 & 70.08$\pm$3.44 & \textbf{79.10$\pm$2.72} \\ \midrule
\multirow{2}{*}{103 more like a drug} & 0 & 36.52$\pm$2.46 & 39.97$\pm$4.32 & 41.49$\pm$2.66 & 32.88+0.99 & 48.65$\pm$3.39 & \textbf{63.30$\pm$1.39} \\
 & 0.1 & 8.81$\pm$0.82 & 14.06$\pm$3.18 & 21.23$\pm$0.96 & 12.38$\pm$3.56 & 19.37$\pm$5.54 & \textbf{30.30$\pm$2.65} \\ \midrule
\multirow{2}{*}{104 less like a drug} & 0 & 58.59$\pm$1.01 & 77.62$\pm$2.80 & \textbf{92.13$\pm$1.29} & 54.96$\pm$1.26 & 70.75$\pm$2.92 & 75.40$\pm$3.22 \\
 & 0.1 & 37.56$\pm$1.76 & 54.22$\pm$3.12 & \textbf{85.83$\pm$1.76} & 29.13$\pm$2.17 & 30.99$\pm$2.66 & 49.60$\pm$2.81 \\ \midrule
\multirow{2}{*}{105 higher permeability} & 0 & 57.74$\pm$0.60 & 59.84$\pm$0.78 & \textbf{84.21$\pm$4.07} & 28.47$\pm$2.76 & 56.56$\pm$1.84 & 81.00$\pm$0.93 \\
 & 10 & 47.51$\pm$1.88 & 50.42$\pm$2.73 & \textbf{79.91$\pm$1.25} & 25.69$\pm$1.20 & 43.08$\pm$2.95 & 66.50$\pm$3.58 \\ \midrule
\multirow{2}{*}{106 lower permeability} & 0 & 34.13$\pm$0.59 & 31.76$\pm$0.97 & 68.49$\pm$2.34 & 47.58$\pm$3.62 & 77.35$\pm$1.98 & \textbf{82.90$\pm$2.04} \\
 & 10 & 26.48$\pm$0.97 & 19.76$\pm$1.31 & 60.46$\pm$0.57 & 38.58$\pm$4.49 & 66.69$\pm$2.74 & \textbf{73.10$\pm$1.29} \\ \midrule
\multirow{2}{*}{107 more hydrogen bond acceptors} & 0 & 54.01$\pm$5.26 & 37.35$\pm$0.79 & 55.18$\pm$3.67 & 10.70$\pm$0.82 & \textbf{95.35$\pm$0.62} & 85.90$\pm$2.40 \\
 & 1 & 27.33$\pm$2.62 & 16.13$\pm$2.87 & 37.40$\pm$1.84 & 7.18$\pm$1.76 & \textbf{72.60$\pm$2.51} & 65.60$\pm$4.32 \\ \midrule
\multirow{2}{*}{108 more hydrogen bond donors} & 0 & 28.55$\pm$0.76 & 60.97$\pm$5.09 & 59.41$\pm$4.07 & 12.77$\pm$4.62 & \textbf{96.54$\pm$1.31} & 95.20$\pm$0.27 \\
 & 1 & 7.69$\pm$0.56 & 32.35$\pm$2.57 & 31.88$\pm$3.22 & 7.15$\pm$2.81 & {76.43$\pm$3.32} & \textbf{79.00$\pm$1.45} \\ \midrule
Average &  & 38.69 &45.54 & 63.31  & 28.96 & 69.01 & \textbf{74.34} \\ \bottomrule
\end{tabular}
}
\end{table}

\begin{table}[!t]
\centering
\caption{{The hit ratio results for 12 multi-objective  molecule editing tasks, along with the average performance across all tasks. For \modelname{}, we present the mean and standard deviation across five random seeds. The best-performing results are highlighted in \textbf{bold}.} The number of experimental replicates $n$=5.}  
\label{tab:response:moledit-multi} 
\scalebox{0.8}{
\begin{tabular}{llcccccc}
\toprule
\multirow{2}{*}{Two Target Property} & \multirow{2}{*}{$\Delta$} & \multicolumn{2}{c}{MoleculeSTM} & \multicolumn{3}{c}{ChatDrug} & \multirow{2}{*}{\modelname{}} \\ \cmidrule(r){3-4} \cmidrule(r){5-7}
 &  & SMILES & Graph & GALACTICA & Llama2 & Turbo &  \\ \midrule
\multirow{2}{*}{\makecell[l]{201 more soluble in water and\\ more hydrogen bond acceptors}} & 0 - 0 & 27.87$\pm$3.86 & 27.43$\pm$3.41 & 39.51$\pm$3.41 & 24.95$\pm$2.55 & \textbf{79.62$\pm$0.64} & 78.50$\pm$3.57 \\
 & 0.5 - 1 & 8.80$\pm$0.04 & 11.10$\pm$1.80 & 26.44$\pm$1.07 & 13.24$\pm$1.17 & 49.64$\pm$2.66 & \textbf{56.20$\pm$3.36} \\ \midrule
\multirow{2}{*}{\makecell[l]{202 less soluble in water and\\ more hydrogen bond acceptors}} & 0 - 0 & 8.55$\pm$2.75 & 8.21$\pm$0.81 & 28.40$\pm$3.11 & 8.91$\pm$1.06 & 51.59$\pm$3.79 & \textbf{61.00$\pm$2.00} \\
 & 0.5 - 1 & 2.93$\pm$0.30 & 0.00$\pm$0.00 & 12.66$\pm$1.40 & 8.27$\pm$1.66 & 24.92$\pm$4.85 & \textbf{39.90$\pm$1.94} \\ \midrule
\multirow{2}{*}{\makecell[l]{203 more soluble in water and\\ more hydrogen bond donors}} & 0 - 0 & 33.51$\pm$4.08 & 49.23$\pm$1.71 & 47.91$\pm$3.33 & 30.66$\pm$2.39 & \textbf{89.34$\pm$0.96} & 88.00$\pm$2.00 \\
 & 0.5 - 1 & 9.98$\pm$1.03 & 23.94$\pm$1.09 & 26.49$\pm$3.37 & 8.17$\pm$3.34 & 53.64$\pm$5.81 & \textbf{68.50$\pm$3.16} \\ \midrule
\multirow{2}{*}{\makecell[l]{204 less soluble in water and\\ more hydrogen bond donors}} & 0 - 0 & 17.03$\pm$2.75 & 14.42$\pm$3.43 & 25.70$\pm$2.07 & 16.30$\pm$4.92 & 39.90$\pm$3.86 & \textbf{76.30$\pm$4.57} \\
 & 0.5 - 1 & 2.59$\pm$1.14 & 3.84$\pm$0.71 & 9.83$\pm$0.85 & 9.04$\pm$1.48 & 24.19$\pm$2.19 & \textbf{47.80$\pm$2.07} \\ \midrule
\multirow{2}{*}{\makecell[l]{205 more soluble in water and\\ higher permeability}} & 0 - 0 & 35.69$\pm$3.19 & 39.74$\pm$2.26 & 56.40$\pm$4.15 & 18.87$\pm$5.02 & 12.85$\pm$2.68 & \textbf{57.10$\pm$3.97} \\
 & 0.5 - 10 & 19.15$\pm$0.73 & 22.66$\pm$1.90 & \textbf{39.22$\pm$0.23} & 15.24$\pm$1.63 & 10.44$\pm$5.75 & 37.60$\pm$2.61 \\ \midrule
\multirow{2}{*}{\makecell[l]{206 more soluble in water and\\ lower permeability}} & 0 - 0 & 44.35$\pm$0.68 & 30.87$\pm$0.62 & 54.87$\pm$0.96 & 41.97$\pm$0.87 & 65.33$\pm$2.16 & \textbf{82.00$\pm$1.83} \\
 & 0.5 - 10 & 28.67$\pm$2.22 & 20.06$\pm$1.26 & 43.91$\pm$1.77 & 35.20$\pm$2.29 & 52.90$\pm$2.23 & \textbf{70.40$\pm$3.74} \\ \midrule
Average &  & 19.93 & 20.96 & 34.28 & 19.24 & 46.20 & \textbf{64.04} \\ \bottomrule
\end{tabular}
}
\end{table}

\begin{table}[t]
\centering
\caption{{Novelty for target protein-based drug design.}}  
\label{tab:response:novel-drug} 
\scalebox{0.75}{  
\begin{tabular}{lccccc}
\toprule
Method  & GPT-3.5 (zero-shot) & GPT-3.5 (5-shot, random) & GPT-3.5 (5-shot, similarity) & DrugGPT & InstructBioMol \\ \midrule
Novelty (\textit{vs} reference) & 100\%    & 100\%   & 100\%    & 98.1\%   & 100\%   \\ \midrule
Novelty (\textit{vs} training) & 99.0\%    & 98.9\%   & 98.0\%    & 91.4\%   & 92.9\%   \\ \bottomrule
\end{tabular}
}
\end{table}

\begin{table}[t]
\centering
\caption{{Novelty for target substrate-based enzyme design.}}  
\label{tab:response:novel-enzyme} 
\scalebox{0.72}{  
\begin{tabular}{lccccc}
\toprule
Method  & GPT-3.5 (zero-shot) & GPT-3.5 (5-shot, random) & GPT-3.5 (5-shot, similarity) & RFdiffusionAA & InstructBioMol \\ \midrule
Novelty (\textit{vs} reference) & 100\%    & 100\%  & 99.1\%    & 100\%   & 99.8\%        \\ \midrule
Novelty (\textit{vs} training) & 100\%    & 99.9\%  & 97.8\%    & 100\%   & 98.0\%        \\ \bottomrule
\end{tabular}
}
\end{table}

\begin{figure*}[t]
\centering
\includegraphics[width=\linewidth]{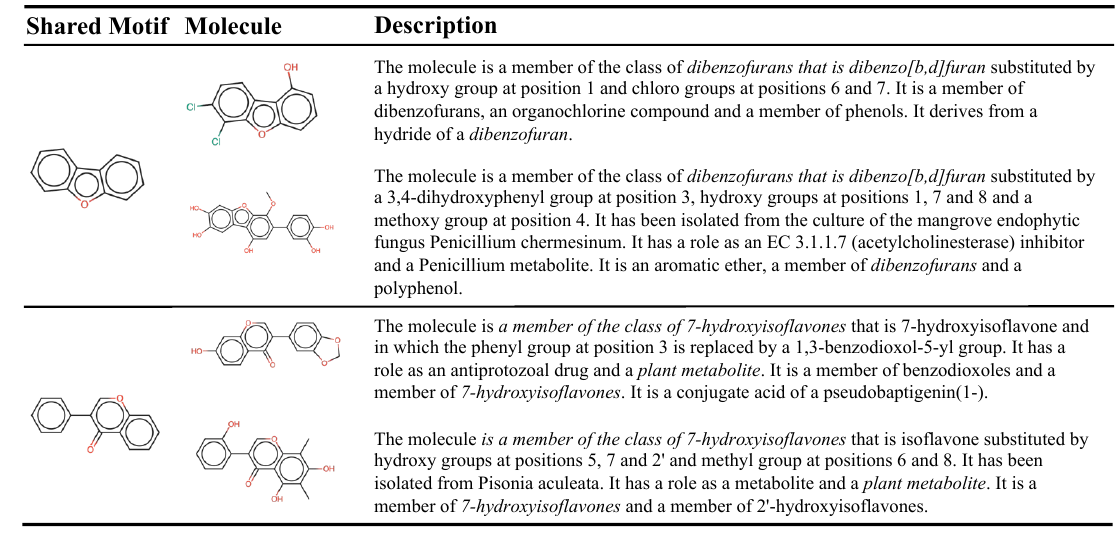}
\caption{{Case analysis of molecular motifs, focusing on shared motifs between molecules as well as their individual structures and corresponding text descriptions. Since the Functional-Class FingerPrint~(FCFP) used as molecular motif information is a hashed vector, we visualize the common substructures between molecules as shared motifs to facilitate a clearer understanding. The common parts in the descriptions are highlighted in \textit{italics}.}}
\label{fig:response:motif-mol}
\end{figure*}

\begin{figure*}[!t]
\centering
\includegraphics[width=\linewidth]{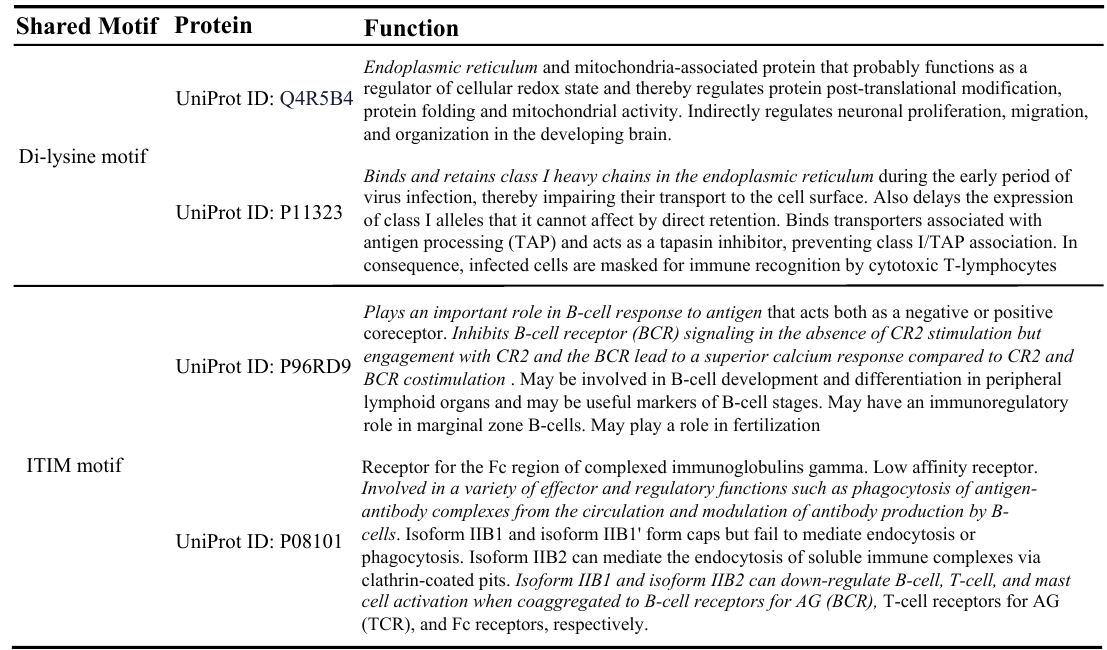}
\caption{{Case analysis of protein motifs, focusing on shared motifs between proteins and corresponding functions. The common parts in the descriptions are highlighted in \textit{italics}.}}
\label{fig:response:motif-prot}
\end{figure*}

\begin{figure*}[!t]
\centering
\includegraphics[width=\linewidth]{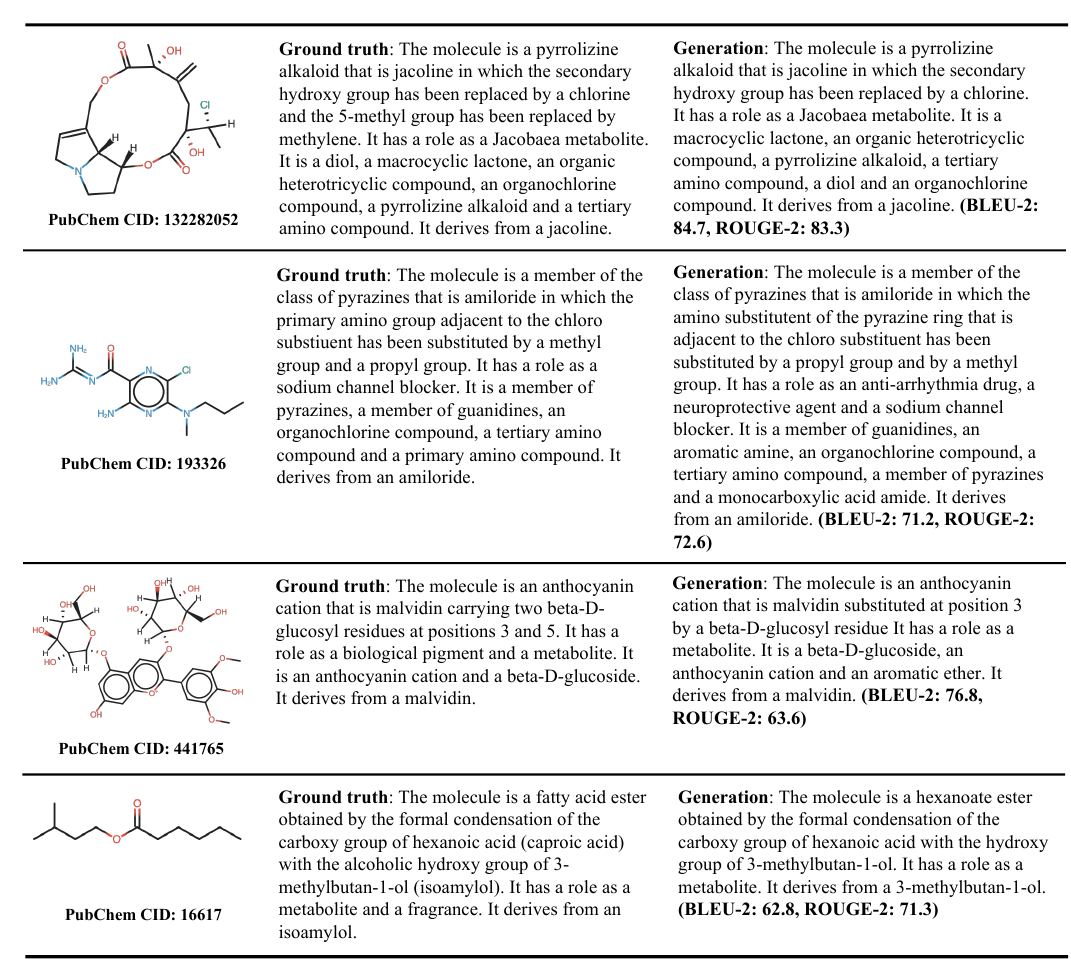}
\caption{Case analysis for molecule captioning task.}
\label{fig:sup-mol2text-demo}
\end{figure*}

\begin{figure*}[!t]
\centering
\includegraphics[width=\linewidth]{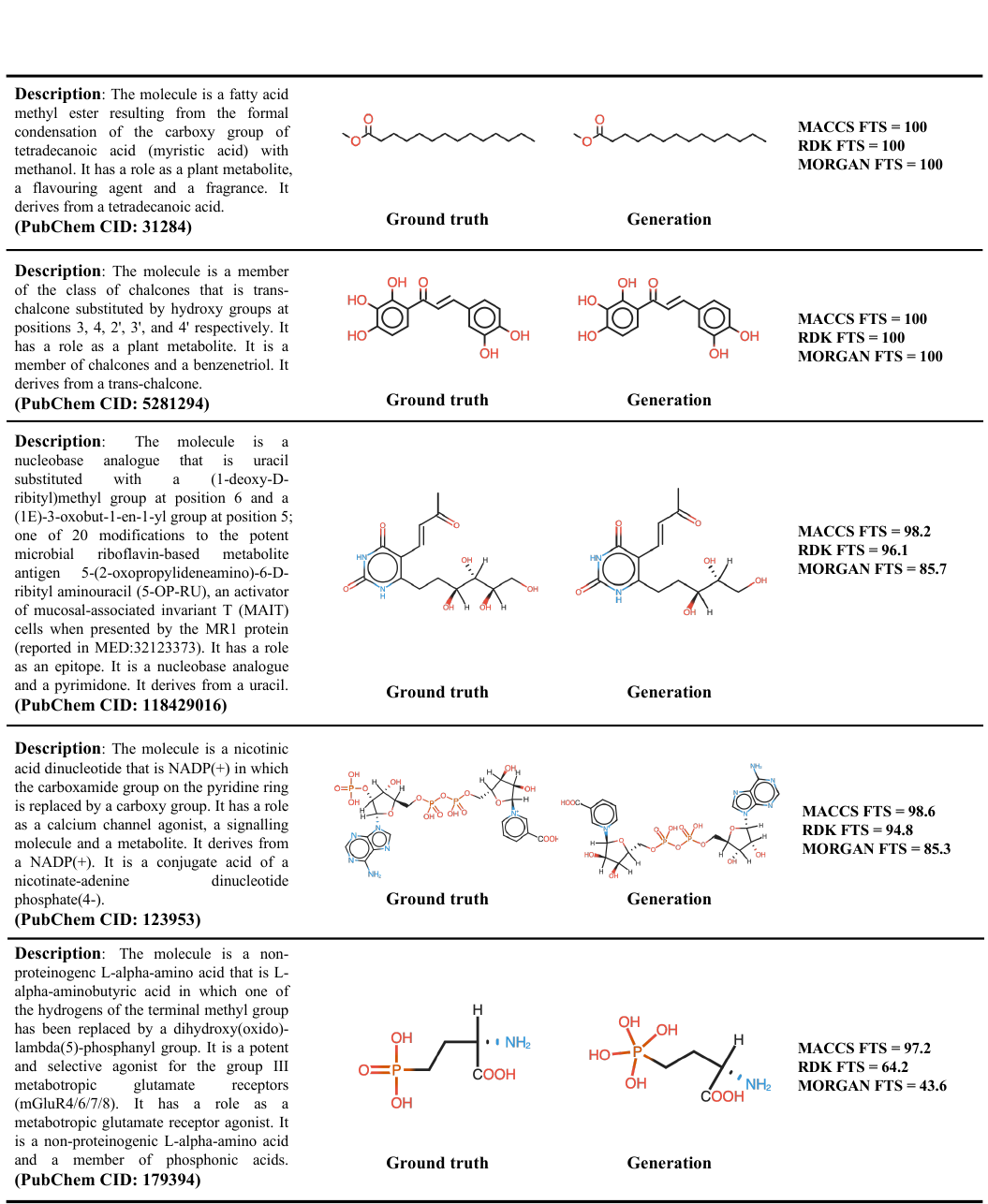}
\caption{Case analysis for description-based molecule generation task.}
\label{fig:sup-text2mol-demo}
\end{figure*}

\begin{table}[t]
\small
\centering
\caption{Case analysis for protein function answering task.}
\scalebox{0.75}{{
\begin{tabular}{lp{8cm}p{8cm}cc}
\toprule
UniProtID & Ground truth & Generation & BLEU-2 & ROURGE-2 \\ \midrule
Q9NRY2 & Component of the SOSS complex, a multiprotein complex that functions downstream of the MRN complex to promote DNA repair and G2/M checkpoint.  The SOSS complex associates with single-stranded DNA at DNA lesions and influences diverse endpoints in the cellular DNA damage response including cell-cycle checkpoint activation, recombinational repair and maintenance of genomic stability.  Required for efficient homologous recombination-dependent repair of double-strand breaks (DSBs) and ATM-dependent signaling pathways. & Component of the SOSS complex, a multiprotein complex that functions downstream of the MRN complex to promote DNA repair and G2/M checkpoint.  The SOSS complex associates with single-stranded DNA at DNA lesions and influences diverse endpoints in the cellular DNA damage response including cell-cycle checkpoint activation, recombinational repair and maintenance of genomic stability.  Required for efficient homologous recombination-dependent repair of double-strand breaks (DSBs). & 92.8 & 96.4 \\ \midrule
P73070 & One of the components of the core complex of photosystem II (PSII), possibly involved in regulating electron flow.  PSII is a light-driven water:plastoquinone oxidoreductase that uses light energy to abstract electrons from H(2)O, generating O(2) and a proton gradient subsequently used for ATP formation.  It consists of a core antenna complex that captures photons, and an electron transfer chain that converts photonic excitation into a charge separation. &  One of the components of the core complex of photosystem II (PSII).  PSII is a light-driven water:plastoquinone oxidoreductase that uses light energy to abstract electrons from H(2)O, generating O(2) and a proton gradient subsequently used for ATP formation.  It consists of a core antenna complex that captures photons, and an electron transfer chain that converts photonic excitation into a charge separation. & 91.7 & 94.1 \\ \midrule
Q9Y2G3 & Catalytic component of a P4-ATPase flippase complex which catalyzes the hydrolysis of ATP coupled to the transport of aminophospholipids, phosphatidylserines (PS) and phosphatidylethanolamines (PE), from the outer to the inner leaflet of intracellular membranes.  May contribute to the maintenance of membrane lipid asymmetry in endosome compartment. & Catalytic component of a P4-ATPase flippase complex which catalyzes the hydrolysis of ATP coupled to the transport of aminophospholipids from the outer to the inner leaflet of various membranes and ensures the maintenance of asymmetric distribution of phospholipids.  Phospholipid translocation seems also to be implicated in vesicle formation and in uptake of lipid signaling molecules.  May also participate in the establishment of the thrombopoietin gradient across the membrane of platelets. & 45.9 & 47.4 \\ \midrule
Q9FY89 & Component of the ESCRT-III complex, which is required for multivesicular bodies (MVBs) formation and sorting of endosomal cargo proteins into MVBs.  The ESCRT-III complex is probably involved in the concentration of MVB cargo. & Probable core component of the endosomal sorting required for transport complex III (ESCRT-III) which is involved in multivesicular bodies (MVBs) formation and sorting of endosomal cargo proteins into MVBs.  MVBs contain intraluminal vesicles (ILVs) that are generated by invagination and scission from the limiting membrane of the endosome and mostly are delivered to lysosomes enabling degradation of membrane proteins, such as stimulated growth factor receptors, lysosomal enzymes and lipids. & 34.8 & 33.0 \\ \midrule
P0CP67 & Responds to activation by environmental stress by phosphorylating downstream targets. & Responds to activation by environmental stress and pro-inflammatory cytokines by phosphorylating a number of transcription factors, primarily components of AP-1 such as c-Jun and ATF2 and thus regulates AP-1 transcriptional activity.  May play a role in the regulation of the circadian clock. & 15.3 & 22.2 \\ \bottomrule
\end{tabular}
}}
\label{tab:sup:protein2text-demo}
\end{table}

\begin{figure*}[!t]
\centering
\includegraphics[width=\linewidth]{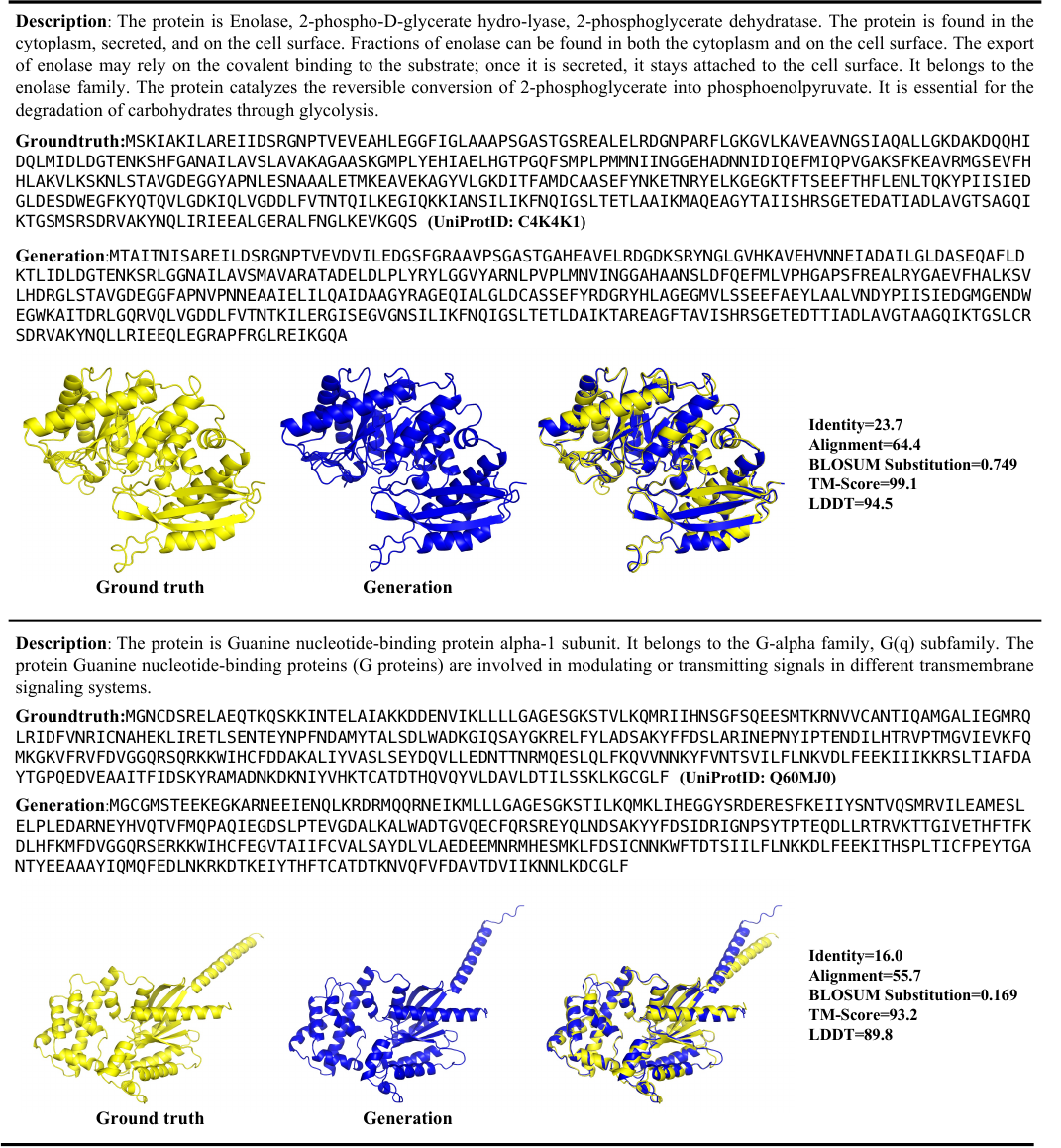}
\caption{Case analysis for description-based protein generation task.}
\label{fig:sup-text2protein-demo}
\end{figure*}

\begin{table}[h]
\centering
\caption{{Detailed experimental results on protein understanding tasks.} The number of experimental replicates $n$=3.}
\scalebox{0.8}{{
\begin{tabular}{lcccccc}
\toprule
 & BLEU-2 & BLEU-4 & ROUGE-1 & ROUGE-2 & ROUGE-L & METEOR \\ \midrule
\multicolumn{7}{l}{\textbf{\textit{Family}}} \\ \midrule
GPT-3.5 (zero-shot) & 1.4$\pm$0.15 & 0.1$\pm$0.01 & 5.5$\pm$0.12 & 0.6$\pm$0.01 & 5.3$\pm$0.14 & 12.0$\pm$0.13 \\
GPT-3.5 (5-shot, random) & 27.4$\pm$3.88 & 6.8$\pm$1.76 & 48.5$\pm$1.14 & 27.3$\pm$0.88 & 47.7$\pm$0.68 & 44.4$\pm$3.05 \\
GPT-3.5 (5-shot, similarity) & 89.0$\pm$0.83 & 86.7$\pm$1.98 & 93.2$\pm$1.30 & 88.9$\pm$1.26 & 91.6$\pm$1.67 & 91.4$\pm$1.81 \\
Mol-Instructions & 1.8$\pm$0.01 & 0.2$\pm$0.05 & 10.7$\pm$0.05 & 0.8$\pm$0.06 & 10.1$\pm$0.01 & 8.5$\pm$0.04 \\
InstructProtein & 8.8$\pm$0.06 & 2.0$\pm$0.02 & 13.5$\pm$0.10 & 0.7$\pm$0.03 & 13.3$\pm$0.09 & 17.8$\pm$0.16 \\
BioT5+ & 5.2$\pm$0.05 & 2.1$\pm$0.02 & 18.0$\pm$0.08 & 3.8$\pm$0.02 & 13.5$\pm$0.01 & 11.0$\pm$0.05 \\
ProtT3 & 13.1$\pm$0.53 & 11.4$\pm$0.52 & 39.8$\pm$0.47 & 35.3$\pm$0.56 & 40.0$\pm$0.58 & 52.4$\pm$0.75 \\
BioMedGPT & 94.6$\pm$0.69 & 92.8$\pm$0.21 & 96.8$\pm$0.58 & 94.1$\pm$0.52 & 96.4$\pm$0.53 & 95.6$\pm$0.32 \\
InstructBioMol & 97.2$\pm$0.43 & 96.5$\pm$0.46 & 98.3$\pm$0.09 & 97.4$\pm$0.11 & 98.3$\pm$0.09 & 98.0$\pm$0.05\\ \midrule
\textbf{\textit{Location}} &  &  &  &  &  &  \\ \midrule
GPT-3.5 (zero-shot) & 0.6$\pm$0.15 & 0.1$\pm$0.03  & 2.6$\pm$0.18  & 0.2$\pm$0.04  & 2.2$\pm$0.04  & 7.2$\pm$0.58  \\
GPT-3.5 (5-shot, random) &  14.1$\pm$3.21 &  8.5$\pm$2.65 & 35.3$\pm$1.02  & 5.8$\pm$0.58  & 34.1$\pm$2.14  & 37.8$\pm$1.61  \\
GPT-3.5 (5-shot, similarity) & 63.4$\pm$1.49  & 56.3$\pm$1.40  & 77.1$\pm$1.14  & 35.1$\pm$0.80  & 75.8$\pm$0.63  & 75.7$\pm$0.92  \\
Mol-Instructions & 0.7$\pm$0.07  & 0.1$\pm$0.05  & 3.4$\pm$0.04  & 0.1$\pm$0.03  & 3.0$\pm$0.04  & 5.0$\pm$0.06  \\
InstructProtein & 1.8$\pm$0.13  & 0.3$\pm$0.09  & 2.1$\pm$0.10  & 0.1$\pm$0.05  & 2.1$\pm$0.13  & 14.4$\pm$0.08  \\
BioT5+ & 1.2$\pm$0.03  & 0.3$\pm$0.02  & 19.4$\pm$0.14  & 4.0$\pm$0.03  & 16.3$\pm$0.11  & 8.0$\pm$0.09  \\
ProtT3 & 6.0$\pm$0.47  & 4.4$\pm$0.54  & 16.8$\pm$0.51  & 10.0$\pm$0.54  & 16.0$\pm$0.36  & 22.9$\pm$0.39  \\
BioMedGPT & 43.8$\pm$0.49  & 38.9$\pm$0.81  & 87.9$\pm$0.96  & 36.3$\pm$0.16  & 87.4$\pm$0.22  & 84.3$\pm$0.62  \\
InstructBioMol & 79.1$\pm$0.65  & 74.6$\pm$0.60  & 93.1$\pm$0.02  & 42.5$\pm$0.08  & 93.0$\pm$0.02  & 90.0$\pm$0.09  \\ \midrule
\textbf{\textit{Name}} &  &  &  &  &  &  \\ \midrule
GPT-3.5 (zero-shot) & 0.6$\pm$0.47  & 0.1$\pm$0.04  & 3.2$\pm$1.55  & 0.1$\pm$0.02  & 3.2$\pm$1.58  & 5.0$\pm$0.97  \\
GPT-3.5 (5-shot, random) & 2.6$\pm$0.35  & 0.7$\pm$0.10  & 7.3$\pm$1.12  & 0.4$\pm$0.10  & 7.2$\pm$1.06  & 8.4$\pm$0.82  \\
GPT-3.5 (5-shot, similarity) & 67.5$\pm$2.28  & 65.2$\pm$1.41  & 70.0$\pm$1.89  & 63.2$\pm$2.02  & 69.6$\pm$0.98  & 67.2$\pm$1.08  \\
Mol-Instructions & 1.0$\pm$0.08 & 0.1$\pm$0.03 & 4.5$\pm$0.05 & 0.1$\pm$0.07 & 4.4$\pm$0.03 & 5.6$\pm$0.07 \\
InstructProtein & 5.7$\pm$0.23   & 2.7$\pm$0.20   & 5.4$\pm$0.12   & 1.2$\pm$0.10   & 5.2$\pm$0.11   & 9.7$\pm$0.21   \\
BioT5+ & 8.4$\pm$0.18  & 4.2$\pm$0.08  & 20.8$\pm$0.14  & 5.8$\pm$0.02  & 16.2$\pm$0.11  & 13.7$\pm$0.06  \\
ProtT3 & 7.1$\pm$0.69  & 3.8$\pm$0.51  & 13.8$\pm$0.15  & 5.4$\pm$0.36  & 12.3$\pm$0.44  & 23.7$\pm$0.56  \\
BioMedGPT & 57.0$\pm$0.38  & 54.0$\pm$0.70  & 74.8$\pm$0.48  & 67.9$\pm$0.41  & 74.4$\pm$0.24  & 75.0$\pm$0.61  \\
InstructBioMol & 81.2$\pm$1.05  & 78.7$\pm$1.17  & 82.3$\pm$0.51  & 77.3$\pm$0.73  & 82.2$\pm$0.49  & 82.9$\pm$0.45  \\ \midrule
\textbf{\textit{Function}} &  &  &  &  &  &  \\ \midrule
GPT-3.5 (zero-shot) & 2.3$\pm$0.74  & 0.3$\pm$0.06  & 10.1$\pm$2.36  & 1.0$\pm$0.03  & 7.4$\pm$1.74  & 12.2$\pm$0.01  \\
GPT-3.5 (5-shot, random) & 4.7$\pm$1.99  & 1.2$\pm$0.51  & 12.9$\pm$2.82  & 1.9$\pm$0.59  & 10.1$\pm$1.98  & 10.6$\pm$2.71  \\
GPT-3.5 (5-shot, similarity) & 57.0$\pm$0.66  & 54.8$\pm$1.47  & 66.0$\pm$1.11  & 62.9$\pm$1.88  & 65.4$\pm$0.87  & 66.0$\pm$1.83  \\
Mol-Instructions & 2.8$\pm$0.07  & 0.5$\pm$0.02  & 10.3$\pm$0.06  & 5.0$\pm$0.02  & 8.1$\pm$0.05  & 6.7$\pm$0.01 \\
InstructProtein & 0.3$\pm$0.16  & 0.1$\pm$0.01  & 4.2$\pm$0.56  & 0.4$\pm$0.12  & 3.8$\pm$0.46  & 5.3$\pm$0.36  \\
BioT5+ & 4.6$\pm$0.02  & 2.0$\pm$0.03  & 20.5$\pm$0.05  & 4.7$\pm$0.03  & 15.2$\pm$0.04  & 12.0$\pm$0.08  \\
ProtT3 & 48.7$\pm$0.45 & 45.0$\pm$0.44 & 60.5$\pm$0.41 & 56.4$\pm$0.88 & 59.4$\pm$0.47 & 64.8$\pm$0.33 \\
BioMedGPT & 47.1$\pm$0.18   & 44.9$\pm$0.09   & 68.6$\pm$0.58   & 62.4$\pm$0.40   & 67.0$\pm$0.58   & 65.3$\pm$0.46   \\
InstructBioMol & 76.9$\pm$0.33  & 75.5$\pm$0.34  & 84.9$\pm$0.24  & 82.7$\pm$0.25  & 84.3$\pm$0.25  & 84.1$\pm$0.29  \\ \bottomrule
\end{tabular}
}}
\label{tab:response:textprot}
\end{table}


\end{document}